\pdfoutput=1

\documentclass[11pt]{article}

\usepackage[final]{acl}
\usepackage{multirow}
\usepackage{times}
\usepackage{latexsym}

\usepackage[T1]{fontenc}

\usepackage[utf8]{inputenc}

\usepackage{microtype}

\usepackage{inconsolata}

\usepackage{graphicx}
\usepackage{amsmath}
\usepackage{subfigure}
\usepackage{colortbl}  
\usepackage{amssymb}
\usepackage{booktabs} 
\usepackage{arydshln}
\usepackage[dvipsnames]{xcolor}

\def\method{P-ALIGN}

\title{Long-Chain Reasoning Distillation via Adaptive Prefix Alignment}

\author{%
  \textbf{Zhenghao Liu}$^{1}$\thanks{~~indicates equal contribution.}\thanks{~~indicates corresponding author.},
  \textbf{Zhuoyang Wu}$^{1}$\footnotemark[1],
  \textbf{Xinze Li}$^{1}$,
  \textbf{Yukun Yan}$^{2}$\footnotemark[2],\\
  \textbf{Shuo Wang}$^{2}$,
  \textbf{Zulong Chen}$^{3}$,
  \textbf{Yu Gu}$^{1}$,
  \textbf{Ge Yu}$^{1}$,
  \textbf{Maosong Sun}$^{2}$\\
  $^{1}$School of Computer Science and Engineering, Northeastern University, China\\
  $^{2}$Department of Computer Science and Technology, Institute for AI, Tsinghua University, China\\
  $^{3}$Alibaba Group, China\\
}

\begin{document}
\maketitle
\begin{abstract}
Large Language Models (LLMs) have demonstrated remarkable reasoning capabilities, particularly in solving complex mathematical problems. Recent studies show that distilling long reasoning trajectories can effectively enhance the reasoning performance of small-scale student models. However, teacher-generated reasoning trajectories are often excessively long and structurally complex, making them difficult for student models to learn. This mismatch leads to a gap between the provided supervision signal and the learning capacity of the student model. To address this challenge, we propose \textbf{P}refix-\textbf{ALIGN}ment distillation (\method), a framework that fully exploits teacher CoTs for distillation through adaptive prefix alignment.
Specifically, \method{} adaptively truncates teacher-generated reasoning trajectories by determining whether the remaining suffix is concise and sufficient to guide the student model. Then, \method{} leverages the teacher-generated prefix to supervise the student model, encouraging effective prefix alignment.
Experiments on multiple mathematical reasoning benchmarks demonstrate that \method{} outperforms all baselines by over 3\%. Further analysis indicates that the prefixes constructed by \method{} provide more effective supervision signals, while avoiding the negative impact of redundant and uncertain reasoning components. All code is available at \url{https://github.com/NEUIR/P-ALIGN}.

\end{abstract}
\section{introduction}
\begin{figure}[t]
    \centering    \small\includegraphics[width=\linewidth]{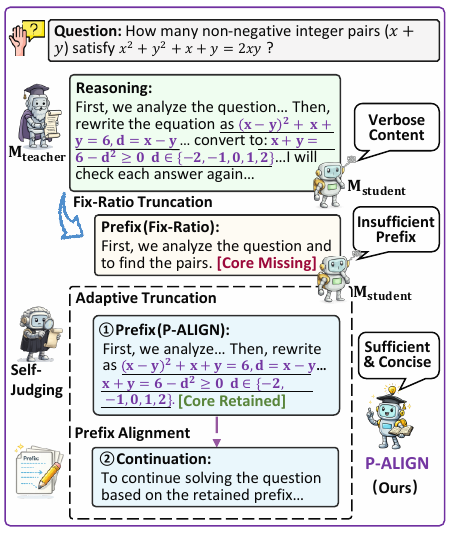}
    \caption{The Framework of Our \method{} Model. Later steps in traces are often more uncertain and harder for the student to learn. \method{} therefore performs adaptively selecting a sufficient prefix and aligning student supervision to it for more effective distillation.}
    \label{fig:intro}
\end{figure}

\noindent Large Language Models (LLMs) have exhibited impressive reasoning capabilities across complex tasks, especially in solving mathematical problems~\cite {brown2020languagemodelsfewshotlearners, zhang2022optopenpretrainedtransformer}. 
By adopting the long-form chain-of-thoughts (CoTs) paradigm~\cite{deepseekai2025deepseekr1incentivizingreasoningcapability}, LLMs can engage in deeper, iterative thinking, which substantially improves their performance on challenging reasoning tasks~\cite{yang2025qwen3technicalreport,chen2025towards}. 
Despite these successes, enabling small-scale models to generate outputs that exhibit such reflective and structured reasoning behavior remains a significant challenge when tackling complex mathematical problems~\cite{luo2025through}.

Recent work has explored distilling reasoning abilities from long-form CoTs to improve the performance of small-scale models on mathematical reasoning benchmarks~\cite{ye2025limoreasoning,deepseekai2025deepseekr1incentivizingreasoningcapability}. In these approaches, small-scale models are treated as student models and trained via Supervised Fine-Tuning (SFT) on full long-form reasoning trajectories generated by Reasoning Language Models (RLMs)~\cite{hsieh2023distillingstepbystepoutperforminglarger}.
However, such intricate and reflective reasoning trajectories often exceed the learning capacity of student models~\cite{li2025smallmodelsstrugglelearn,anonymous2025morale}, making it difficult for them to internalize the underlying reasoning patterns and, in some cases, even leading to degraded reasoning performance~\cite{luo2025through}. To address this issue, several recent studies attempt to refine teacher-generated trajectories using LLM-based refinement techniques to reduce redundancy and improve the compatibility of distilled reasoning with student models~\cite{wu2025enhancing,zeng2025pruning}. Nevertheless, refinement-based methods introduce an additional and weakly constrained transformation stage, where the refined trajectories are not explicitly optimized to match the capacity of the student model.
To better exploit teacher-generated reasoning trajectories, existing works~\cite{UPFT,chen2025datacentricsamplecentricenhancingllm} incorporate prefixes into the SFT process, as these prefixes help preserve structural knowledge while mitigating uncertainty and redundancy that tend to emerge in later reasoning steps. However, as shown in Figure~\ref{fig:intro}, using a fixed ratio for prefix truncation renders this strategy suboptimal: prefixes may still fail to retain sufficient information for complex tasks.


In this paper, we introduce the Adaptive \textbf{P}refix-\textbf{ALIGN}ment reasoning distillation method (\method), which adaptively truncates prefixes from teacher-generated reasoning trajectories and optimizes student models to align with the retained teacher prefixes. Specifically, \method{} enables the student model to adaptively assess the sufficiency of a truncated prefix and identify a minimal sufficient prefix boundary via binary search, ensuring that the retained prefix is both effective and concise. Building on this adaptive prefix truncation strategy, \method{} further performs prefix-based alignment by treating the retained teacher prefix as prior reasoning context and guiding the student to generate a complete reasoning trajectory, which serves as the supervision signal for SFT.

Our experimental results demonstrate that \method{} consistently outperforms all baselines across a range of mathematical tasks, highlighting its effectiveness in distilling long-form reasoning into student models.
Further analysis reveals that reasoning content from later steps introduces increased uncertainty into student models, which in turn degrades distillation performance. Therefore, to fully exploit the teacher-generated reasoning trajectories, \method{} adaptively applies prefix truncation to the reasoning trajectories, retaining more informative content while reducing unnecessary uncertainty and bias when guiding the student model on problems of varying difficulty.
In addition, the prefix alignment strategy in \method{} further demonstrates its effectiveness under the SFT setting by enabling the student model to generate complete reasoning chains autonomously. This design avoids overfitting to incomplete or overly short reasoning patterns, a common issue in prefix-based SFT methods~\cite{UPFT}.

\section{Reated Work}
Large Language Models (LLMs) have demonstrated strong capabilities in mathematical reasoning~\cite{cobbe2021trainingverifierssolvemath,hendrycks2021measuringmathematicalproblemsolving}.
Chain-of-Thought (CoT)~\cite{wei2022chain} elicits step-by-step reasoning and has been shown to substantially improve performance on mathematical reasoning benchmarks~\cite{li2023query,qin2023cross,luo2023wizardmath}.
To enhance small-scale student models, prior work commonly leverages a stronger LLM as a teacher to generate long-form reasoning trajectories and uses these trajectories as supervision signals to fine-tune student models~\cite{deepseekai2025deepseekr1incentivizingreasoningcapability,yang2025qwen3technicalreport}.
However, the uncertainty and redundancy of teacher-generated long-form CoTs often exceed the capacity of student models, making it challenging to faithfully distill such reasoning capabilities into them~\cite{chen2025towards,li2025smallmodelsstrugglelearn}.

To mitigate the challenges of distilling long-form reasoning trajectories, recent work has increasingly focused on constructing more efficient and higher-quality SFT data. One line of research investigates quality-based data selection, where sophisticated criteria are employed to identify a small set of high-quality examples for training~\cite{ye2025limoreasoning,muennighoff2025s1simpletesttimescaling}. Another line of work refines teacher-generated reasoning trajectories through prompting strategies or structural editing, producing shorter yet still effective trajectories~\cite{xu2025chain,wu2025enhancing,jin2025recut}. However, these selection and refinement approaches typically involve multiple sequential stages, such as iterative improvements~\cite{zelikman2022star} and verification, to refine or select high-quality long-form reasoning trajectories.


Unlike these sophisticated CoT selection or refinement methods, recent studies have proposed unsupervised approaches to fully exploit the effectiveness of teacher-generated reasoning trajectories.
Specifically, \citet{UPFT} observe that the prefixes of reasoning trajectories are typically consistent; they then extract such prefixes and mix them with full CoTs as supervision during SFT. Subsequent studies~\cite{chen2025datacentricsamplecentricenhancingllm,sun2025begunhalfdonereinforcement} further explore the use of randomly truncated prefixes from teacher-generated reasoning trajectories to either enhance reasoning performance or improve trajectory sampling efficiency in RL training. However, employing a fixed truncation ratio lacks adaptability across problem difficulties, often failing to preserve reasoning information that is both sufficient for complex tasks and concise for simpler ones.

\section{Methodology}
\begin{figure*}[t]
    \centering
    \includegraphics[width=\linewidth]{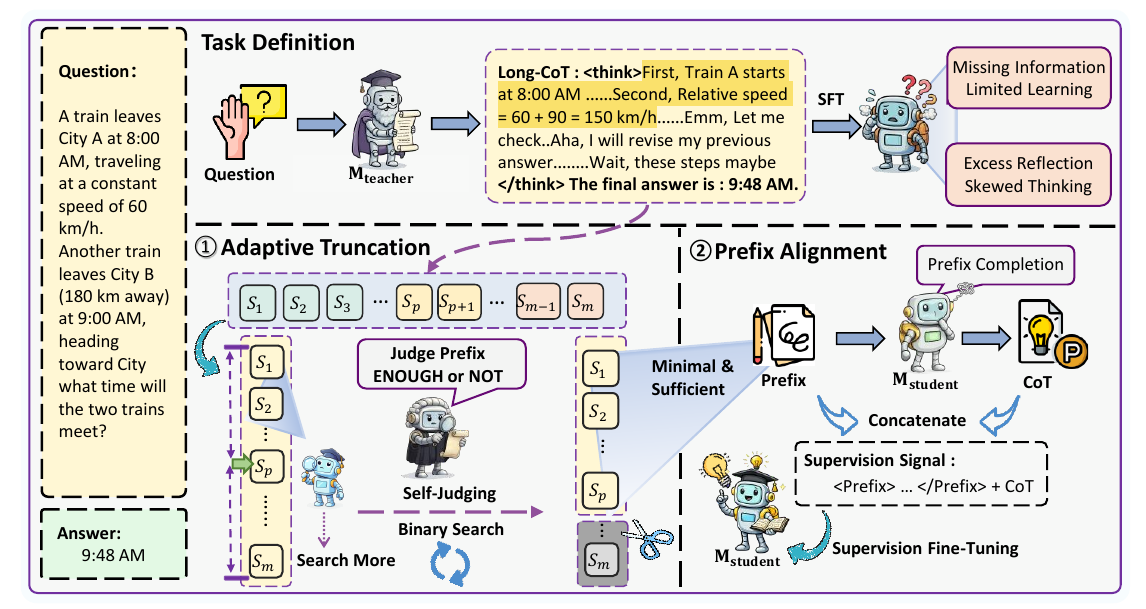}
    \caption{Illustration of \method{} Model.}
    \label{fig:model}
\end{figure*}

In this section, we present \textbf{P}refix-\textbf{ALIGN}ment distillation (\method{}), a framework designed to distill long-chain reasoning capabilities into small-scaled models efficiently. As shown in Figure~\ref{fig:model}, we first describe the standard supervised fine-tuning (SFT) for reasoning distillation, along with its truncated-prefix SFT variant (Sec.~\ref{sec:3.1}). We then introduce our \method{} model (Sec.~\ref{sec:3.2}), in which the student model evaluates reasoning trajectories and uses binary search to truncate them into minimal, sufficient prefixes aligned with its needs to solve the given problem. Then \method{} treats the retained prefixes as prior reasoning context to guide prefix-aligned generation of full reasoning sequences for effective supervision.

\subsection{Distilling Long-Chain Reasoning via Supervised Fine-Tuning}\label{sec:3.1}
Given a mathematical problem $q$, a Large Language Model (LLM) is typically prompted with a problem-solving instruction ($\text{Instruct}_\text{QA}$) to generate a complete solution. In reasoning distillation, we treat the small-scale model as the student model $\mathcal{M}_\text{student}$ and employ different Supervised Fine-Tuning (SFT) strategies to optimize it using signals derived from a stronger teacher model.

\textbf{Distillation from Long-Form CoT.}
To distill the reasoning capability of a teacher model $\mathcal{M}_\text{teacher}$ into a student model $\mathcal{M}_\text{student}$, existing approaches~\cite{wang2023making,yin2025marcoo1v2wideningdistillation,deepseekai2025deepseekr1incentivizingreasoningcapability} typically use the reasoning trajectories generated by $\mathcal{M}_\text{teacher}$ as supervision signals for SFT. Specifically, given a question $q$, we adopt a Reasoning Language Model (RLM), such as DeepSeek-R1~\cite{deepseekai2025deepseekr1incentivizingreasoningcapability}, as the teacher model and prompt it to produce a long-form reasoning response $R$:
\begin{equation}\small\label{eq:error}
R = \mathcal{M}_\text{teacher}(\text{Instruct}_\text{QA}(q)).
\end{equation}
We then collect the query–response pairs as the training dataset $\mathcal{D} = \{(q^1, R^1), \dots, (q^n, R^n)\}$, which consists of $n$ examples.
The training objective minimizes the negative log-likelihood of the teacher-generated reasoning trajectory $R^i$:
\begin{equation}\label{eq:longcot-sft}
\small
\mathcal{J} = -\sum_{i=1}^{n} \sum_{t=1}^{|R^i|}\log P(R_t^i|R_{<t}^i,\text{Instruct}_\text{QA}(q^i);\mathcal{M}_\text{student}),
\end{equation}
where $|R^i|$ denotes the token number of the generated reasoning trajectory $R^i$. Although distilling long-form CoTs from the teacher model substantially improves LLM problem-solving accuracy~\cite{hsieh2023distillingstepbystepoutperforminglarger}, the student model may struggle to effectively learn from these long-form reasoning trajectories~\cite{luo2023empirical,gudibande2023false}. This limitation has motivated recent efforts toward constructing higher-quality SFT datasets~\cite{wettig2024qurating}.

\textbf{SFT with Truncated Prefixes.}
To further improve the quality of supervision signals, recent studies~\cite{UPFT,chen2025datacentricsamplecentricenhancingllm} demonstrate that incorporating truncated prefixes during SFT can enhance model performance, since different solution paths often share a common initial reasoning trajectory. Accordingly, they construct a prefix-truncated dataset $\mathcal{D}_\text{Prefix}$ to encourage the student model to better learn from the early stages of reasoning trajectories, which is then mixed with the original training dataset $\mathcal{D}$ for SFT.

To construct the prefix-truncated dataset $\mathcal{D}_\text{Prefix}$, the $\mathcal{M}_\text{teacher}$-generated reasoning trajectory $R^{i}$ is truncated using the function $\text{Truncate}(\cdot)$:
\begin{equation}\small
\tilde{R}^i = \text{Truncate}(R^i, \lambda \cdot |R^i|),
\end{equation}
where $\lambda \in (0,1)$ controls the truncation ratio and $\text{Truncate}(\cdot)$ retains the first $\lambda\cdot|R^i|$ tokens of $R^i$. The resulting prefix $\tilde{R}^i$ is directly used as the supervision signal for training. We then construct the prefix SFT dataset $\mathcal{D}_\text{Prefix} = \{(q^1, \tilde{{R}^1}), \dots, (q^n, \tilde{{R}^n})\}$ by pairing each input query $q^i$ with its corresponding truncated reasoning prefix. However, such ratio-based truncation strategies may still discard necessary information from the reasoning trajectories, potentially leading to incomplete or incorrect reasoning outcomes.

\subsection{Adaptive Prefix Alignment based Reasoning Capability Distillation}\label{sec:3.2}
To further enhance the distillation of the reasoning capability from $\mathcal{M}_\text{teacher}$, \method{} introduces an adaptive prefix alignment strategy, which encourages the student model $\mathcal{M}_\text{student}$ to dynamically truncate prefixes from reasoning trajectories. Specifically, \method{} prompts $\mathcal{M}_\text{student}$ to assess the sufficiency of a prefix for truncation, while ensuring minimality so as to avoid the introduction of redundant or noisy information. Subsequently, the student model $\mathcal{M}_\text{student}$ is optimized via SFT to align with the truncated prefixes generated by the teacher model $\mathcal{M}_\text{teacher}$.

\subsubsection{Efficient Minimal Prefix Truncation via Binary Search}\label{sec:3.2.1}
To adaptively truncate prefixes from reasoning trajectories, \method{} prompts the student model $\mathcal{M}_\text{student}$ to progressively read prefix sentences from a teacher-generated reasoning trajectory $R^i$ and determine whether the truncated prefix contains sufficient information to solve the problem, while maintaining a minimal length. To enable efficient identification of such a prefix, we also employ a binary search algorithm to locate a tailored truncation point.

\textbf{Prefix Evaluation by Self-Judging.} For the $i$-th reasoning trajectory $R^i$ generated by the teacher model $\mathcal{M}_\text{teacher}$, \method{} first segments $R^i$ into $m$ sentences:
\begin{equation}\small
\label{eq:sent_units}
R^i = \{r_1^i, r_2^i, \dots, r_m^i\}.
\end{equation}
These sentences serve as the basic units for prefix truncation, rather than individual tokens~\cite{chen2025datacentricsamplecentricenhancingllm}, in order to preserve the semantic completeness of the truncated prefixes.
We then prompt the student model $\mathcal{M}_\text{student}$ to self-evaluate whether the current truncated prefix $\tilde{R}^i$ contains sufficient information to solve the problem $q^i$:
\begin{equation}\label{eq:is_enough}
\small
L = \mathcal{M}_\text{student}(\text{Instruct}_{\text{Eval}}(q^i,\tilde{R}^i)),
\end{equation}
where $\text{Instruct}_{\text{Eval}}$ denotes an evaluation instruction that guides $\mathcal{M}_\text{student}$ to make a sufficiency judgement.
The judgment label $L \in \{\texttt{ENOUGH},\texttt{NOT\_ENOUGH} \}$ represents the predicted sufficiency of the current prefix $\tilde{R}^i$.

\textbf{Binary Search for Efficient Truncation.} To efficiently identify the minimal sufficient truncation point for $q^i$, we perform binary search over the sentence sequence $R^i = \{r_1^i, r_2^i, \dots, r_m^i\}$ to locate the shortest prefix that satisfies the sufficiency criterion.
In the binary search procedure, the search boundaries $\ell$ and $r$ are initialized as $\ell = 1$ and $r = m$. The truncation point $p$ is selected as the midpoint of the current interval:
\begin{equation}\small
\label{eq:midpoint}
p = \left\lfloor \frac{\ell + r}{2} \right\rfloor.
\end{equation}
We truncate $R^i$ to its first $p$ sentences to obtain the candidate prefix $\tilde{R}^i = R^i_{1:p}$.
If the evaluation result is \texttt{ENOUGH}, the current truncation point $p$ is deemed feasible, and we continue searching for a shorter sufficient prefix by updating the right boundary to $r = p$. Conversely, if the label is \texttt{NOT\_ENOUGH}, the prefix lacks sufficient information, and we extend the search to the right half by setting $\ell = p + 1$. Formally, the boundary update rule is defined as:
\begin{equation}\small
\label{eq:bs_update}
(\ell, r) =
\begin{cases}
(\ell, p), & \text{if } L=\texttt{ENOUGH},\\
(p+1, r), & \text{if } L=\texttt{NOT\_ENOUGH}.
\end{cases}
\end{equation}
This binary search process is repeated on the updated interval $[\ell, r]$ by recomputing the midpoint in Eq.~\ref{eq:midpoint} and adjusting the boundaries according to Eq.~\ref{eq:bs_update}. The search terminates when $\ell = r$, yielding the final truncation point $p^*$ that is equal to $\ell$. The resulting minimal prefix $\tilde{R}^i=R^i_{1:p^*}$ thus consists of the first $p^*$ sentences of $R^i$. For clarity, we provide a complete procedure of this binary search in Appendix~\ref{app:binary_search_case}.

\subsubsection{Prefix-based Alignment for Effective Supervised Fine-Tuning}\label{sec:3.2.2}
To better align the student model with the retained prefixes $\tilde{R}^i$ introduced in Sec.~\ref{sec:3.2.1}, we first collect all query-prefix pairs into a dataset $\mathcal{D}_{\text{Prefix}}=\{(q^1,\tilde{R}^1),\dots,(q^n,\tilde{R}^n)\}$. Based on this dataset, we further complete each retained prefix into a full reasoning trajectory, thereby constructing the alignment dataset ${\mathcal{D}}_{\text{align}}$ for training the student model $\mathcal{M}_\text{student}$.

Specifically, for each pair $(q^i,\tilde{R}^i)$, an instruction $\text{Instruct}_{\text{Align}}$ prompts the student model $\mathcal{M}_\text{student}$ to condition on $\tilde{R}^i$ as prior reasoning context and then continue the reasoning process to generate a complete reasoning trajectory that is consistent with the given prefix:
\begin{equation}\small
\label{eq:prefix_align_gen}
y^i = \mathcal{M}_\text{student}(\text{Instruct}_{\text{Align}}(q^i,\tilde{R}^i)).
\end{equation}
To enforce prefix-based alignment during SFT, we concatenate the retained prefix $\tilde{R}^i$ with the student-generated continuation $y^i$ as the supervision signal.
To ensure the correctness of the constructed supervision signals, we conduct the $\mathcal{D}_{\text{Align}}$ dataset by retaining only those samples whose final answers match the ground-truth answer $a^i_*$:
\begin{equation}\small\label{eq:filter}
{\mathcal{D}}_{\text{align}} = \{ (q^{i}, \tilde{R}^{i}\oplus y^{i}) \mid 1 \le {i} \le n,\ \text{Ans}(y^{i}) = a^i_*\},
\end{equation}
where $\text{Ans}(\cdot)$ denotes an answer extraction function. The resulting dataset $\mathcal{D}_\text{align}$ is then used for fine-tuning the student model $\mathcal{M}_\text{student}$ using Eq.~\ref{eq:longcot-sft}, enabling the student model to align its reasoning prefixes with those of the teacher model.

\begin{table*}[t]
\centering
\small
\resizebox{\textwidth}{!}{
\begin{tabular}{l| r r r r r r r r r r}
\hline
\multicolumn{1}{l|}{\textbf{}} &
\multicolumn{2}{c}{\textbf{AIME25}} &
\multicolumn{2}{c}{\textbf{AIME24}} &
\multicolumn{2}{c}{\textbf{AMC12}} &
\multicolumn{2}{c}{\textbf{MATH500}} &
\multicolumn{2}{c}{\textbf{Avg.}} \\

& Pass@1 & Pass@3 & Pass@1 & Pass@3 & Pass@1 & Pass@3 & Pass@1 & Pass@3 & Pass@1 & Pass@3 \\
\hline

\multicolumn{11}{c}{\cellcolor{gray!8}{ \textbf{Qwen2.5-7B-Instruct}}} \\ \hline

Zero-Shot  & 6.67  & 10.00  & 10.00  & 20.00  & 40.76  & 49.70  & 69.80  & 84.80  & 31.81  & 41.13  \\
SFT (Label)  &3.33 & 6.67 & 3.33 & 10.00 & 32.92 & 34.62 & 62.20 & 71.00 & 25.45 & 30.57 \\
SFT (Long-CoT)       & 10.00 & 20.00  & 10.00  & 26.67  & 48.19  & 56.63  & 75.60  & \textbf{86.60}  & 35.95  & 47.68  \\
UPFT~\shortcite{UPFT}  & 13.33 & 20.00 & 13.33 & 23.33 & 47.58 & 59.04 & 74.80 & 84.60 & 37.26 & 46.74  \\

\method{}  & \textbf{16.67} & \textbf{26.67} & \textbf{16.67} & \textbf{26.67} & \textbf{49.40} & \textbf{63.86} & \textbf{75.80} & 85.20 & \textbf{39.64} & \textbf{50.60} \\
\hline

\multicolumn{11}{c}{\cellcolor{gray!8}{ \textbf{Qwen3-8B}}} \\ \hline

Zero-Shot  & 20.00 & 23.33 & 26.67 & 36.67 & 59.84 & 71.08 & 83.60 & 91.20 & 47.53 & 55.57 \\
SFT (Label)  &10.00 & 13.33 & 10.00 & 10.00 & 48.51 & 59.04 & 73.20 & 84.60 & 35.43 & 41.74 \\
SFT (Long-CoT) & 23.33 & 30.00 & 26.67 & \textbf{40.00} & 58.81 & 67.04 & 84.40 & 90.80 & 48.30 & 56.96 \\
UPFT~\shortcite{UPFT}  & 23.33 & 26.67 & 23.33 & 36.67 & 61.74 & 75.38 & \textbf{85.60} & 92.40 & 48.50 & 57.78  \\

\method{} &\textbf{23.33} & \textbf{33.33} & \textbf{30.00} & 36.67 & \textbf{66.27} & \textbf{77.11} & 84.80 & \textbf{92.80} & \textbf{51.10} & \textbf{59.98}  \\
\hline
\end{tabular}}
\caption{Overall Performance.}
\label{tab:overall}
\end{table*}

\section{Experimental Methodology}
This section first introduces the datasets, baselines, and evaluation metrics, and then details the implementation settings used in our experiments.

\textbf{Dataset.} We construct the training data using the training split of the s1K-1.1 dataset~\cite{muennighoff2025s1simpletesttimescaling}, which contains high-quality solution traces generated by the DeepSeek-R1 model~\cite{deepseekai2025deepseekr1incentivizingreasoningcapability}. For evaluation, we consider four mathematical problem datasets that span a wide range of difficulty levels. MATH500~\citep{godahewa2021monashtimeseriesforecasting} is a benchmark composed of competition-level mathematics problems with varying degrees of complexity. AIME~\citep{AIME} and AMC~\cite{AMC12} assess mathematical problem-solving capabilities across arithmetic, algebra, counting, geometry, number theory, probability, and other topics in secondary school mathematics.

\textbf{Baselines.}
We compare \method{} against both zero-shot and SFT-based baselines.
In the zero-shot setting, the model is prompted with the question and directly generates a solution without task-specific fine-tuning. We further include three SFT baselines: SFT (Label), SFT (Long-CoT)~\citep{deepseekai2025deepseekr1incentivizingreasoningcapability}, and UPFT~\cite{UPFT}. SFT (Label) trains the student model using ground-truth labels provided in the dataset. Following prior work~\citep{wang2024learning}, SFT (Long-CoT) fine-tunes the student model using full-length reasoning traces produced by the teacher model. In addition, UPFT selects prefixes based on self-consistency and conducts training on truncated prefixes, using the first 32 tokens in accordance with the original experimental setup.

\textbf{Evaluation Metric.}
Following prior work~\citep{reich2023overcome,zhang2024wrong}, we evaluate model performance using Pass@1 and Pass@3. Pass@1 measures the accuracy of a single generated answer, while Pass@3 denotes the proportion of problems for which at least one of three sampled outputs matches the ground-truth answer.

\textbf{Implementation Details.}
All experiments employ Qwen2.5-7B-Instruct~\citep{qwen2025qwen25technicalreport} and Qwen3-8B~\citep{yang2025qwen3technicalreport} as student models, with DeepSeek-R1~\cite{deepseekai2025deepseekr1incentivizingreasoningcapability} serving as the teacher model. We train the models for 3 epochs with a learning rate of $5\times10^{-5}$ and adopt LoRA~\citep{hulora} for parameter-efficient fine-tuning. Our \method{} is implemented based on TRL\footnote{\url{https://github.com/huggingface/trl}} and LLaMA Factory\footnote{\url{https://github.com/hiyouga/LLaMA-Factory}}. Additional experimental details are provided in Appendix~\ref{app:data-statistics}, and the prompt templates are presented in Appendix~\ref{app:prompt-price}.
\section{Evaluation Result}
\begin{table*}[t]
\centering
\small
\resizebox{\textwidth}{!}{
\begin{tabular}{l|rrrrrrrrrr}
\hline
{\textbf{}} & \multicolumn{2}{c}{\textbf{AIME25}} & \multicolumn{2}{c}
{\textbf{AIME24}}& \multicolumn{2}{c}
{\textbf{AMC12}} & \multicolumn{2}{c}{\textbf{MATH500}} & \multicolumn{2}{c}{\textbf{Avg.}} \\ 
~ &Pass@1 & Pass@3 & Pass@1 & Pass@3 & Pass@1 & Pass@3 & Pass@1 & Pass@3 & Pass@1 & Pass@3  \\ \hline

\multicolumn{11}{c}{\cellcolor{gray!8}{%
   \textbf{Qwen2.5-7B-Instruct}}
} \\ \hline

SFT (Long-CoT) & 10.00 & 20.00  & 10.00  & 26.67  & 48.19  & 56.63  & 75.60  & \textbf{86.60}  & 35.95  & 47.68  \\

\method{} & \textbf{16.67} & \textbf{26.67} & \textbf{16.67} & \textbf{26.67} & \textbf{49.40} & 63.86 & \textbf{75.80} & 85.20 & \textbf{39.64} & \textbf{50.60} \\
SFT w/ Teacher Prefix & 13.33 & 20.00 & 13.33 & 23.33 & 46.71 & 54.22 & 70.20 & 82.80 & 35.89 & 45.09 \\ 
SFT w/ Student CoT & 13.33 & 16.67 & 13.33 & 20.00 & 40.55 & 56.63 & 71.60 & 83.80 & 34.70 & 44.28\\
\cdashline{1-11}
\method{} (InfoGain) & 13.33 & 20.00  & 10.00  & 20.00  & 42.17  & 55.90  & 73.80  & 85.40  & 34.83  & 45.33  \\
\method{} w/o Adaptive Read & 13.33 & 16.67 & 16.67 & 23.33 & 47.53 & 60.19 & 73.60 & 84.80 & 37.78 & 46.25 \\
\method{} w/o Binary Search & 16.67 & 23.33 & 16.67 & 26.67 & 49.26 & \textbf{64.07} & 75.00 & 85.20 & 39.40 & 49.82 \\

\hline
\multicolumn{11}{c}{\cellcolor{gray!8}{%
   \textbf{Qwen3-8B}}
}  \\ \hline
SFT (Long-CoT) & 23.33 & 30.00 & 26.67 & 40.00 & 58.81 & 67.04 & 84.40 & 90.80 & 48.30 & 56.96  \\

\method{} &23.33 & \textbf{33.33} & \textbf{30.00} & 36.67 & \textbf{66.27} & \textbf{77.11} & 84.80 & \textbf{92.80} & \textbf{51.10} & \textbf{59.98} \\ 
SFT w/ Teacher Prefix & 16.67 & 30.00 & 26.67 & 33.33 & 60.24 & 72.29 & 83.60 & 92.20 & 46.80 & 56.96  \\
SFT w/ Student CoT & 16.67 & 30.00 & 23.33 & 30.00 & 60.04 & 71.08 & 84.60 & 91.20 & 46.16 & 55.57\\
\cdashline{1-11}
\method{} (InfoGain) & 23.33 & 26.67 & 23.33 & 33.33 & 63.58 & 73.21 & 84.20 & 88.40 & 48.61 & 55.40  \\
\method{} w/o Adaptive Read & 23.33 & 26.67 & 26.67 & \textbf{40.00} & 62.75 & 75.64 & 82.20 & 92.80 & 48.74 & 58.36 \\
\method{} w/o Binary Search & \textbf{26.67} & 26.67 & 26.67 & 40.00 & 62.63 & 76.34 & \textbf{85.00} & 91.40 & 50.24 & 58.60 \\
 \hline

\end{tabular}}
\caption{Ablation Study.}
\label{tab:ablation}
\end{table*}

In this section, we first evaluate the overall performance of \method{} through the main experience. Next, we conduct ablation studies to examine the contribution of different components in \method{}. Furthermore, we analyze the effectiveness of distilled models under different prefix truncation strategies. Finally, we investigate positional effects in long-form reasoning trajectories, aiming to analyse how prefix-based designs contribute to improved model performance. The case study is conducted in Appendix~\ref{app:case-study}.

\subsection{Overall Performance}
As shown in Table~\ref{tab:overall}, we compare \method{} with several baseline models across different mathematical reasoning tasks.

The evaluation results indicate that \method{} consistently outperforms all baselines, highlighting the effectiveness of our prefix-alignment framework for reasoning distillation. Notably, \method{} achieves consistent improvements across different backbone models, demonstrating its generalization ability.
Among different SFT strategies, SFT (Label) performs substantially worse than the zero-shot model, suggesting that supervising the student model solely with annotated ground-truth targets may lead to overfitting and consequently degrade its reasoning capability. In contrast, SFT (Long-CoT) consistently improves the reasoning performance of student models, as teacher-generated long-form reasoning trajectories provide richer and more informative patterns for imitation.
Building upon the distillation of teacher-generated long-form reasoning trajectories, \method{} further constructs higher-quality supervision signals by adaptively truncating a minimal sufficient prefix through student self-judging and performing prefix-based alignment for distillation, yielding an additional 3\% improvement over SFT (Long-CoT).
Compared with UPFT, \method{} achieves an improvement of more than 2\%, demonstrating the crucial role of adaptive prefix alignment, rather than relying on brute-force prefix truncation with a fixed token budget for SFT.
In addition, we report results for student models at different scales in Appendix~\ref{app:other_student_models}.

\subsection{Ablation Study}
This subsection conducts ablation studies to investigate the contributions of different components in our \method{} model. 

As shown in Table~\ref{tab:ablation}, we first evaluate two variants, SFT w/ Teacher Prefix and SFT w/ Student CoT, to analyze the impact of different supervision signals. Specifically, SFT w/ Teacher Prefix only leverages the teacher-generated prefix, while SFT w/ Student CoT relies solely on student-generated reasoning conditioned on the teacher-generated prefix. 
We then conduct three additional variants to examine the effectiveness of the prefix truncation strategy in \method{}. The \method{} (InfoGain) model adopts InfoGain~\cite{wang2025infogain} as the criterion for prefix truncation. \method{} w/o Binary Search removes the binary search procedure for identifying the truncation point. In contrast, \method{} w/o Adaptive Read directly feeds the entire teacher-generated trajectory to the student model without adaptive truncation.

Compared with \method{}, both SFT w/ Teacher Prefix and SFT w/ Student CoT decrease the answering accuracy by more than 3\%, highlighting the effectiveness of incorporating both supervision components. Each component contributes to constructing higher-quality supervision, which helps distill the teacher's reasoning patterns into the student model. The main reason may lie in that the teacher prefix guides the student model to align with the teacher's reasoning process, while the student-generated CoT provides more complete and prefix-guided reasoning trajectories for SFT.
We then ablate different prefix truncation strategies to further verify the effectiveness of our truncation procedure. The InfoGain-based variant determines prefix boundaries based on changes in information entropy~\cite{wang2025infogain}. Although it improves over the vanilla LLM, it underperforms SFT (Long-CoT), demonstrating the advantage of using student self-judging to assess prefix sufficiency.
Next, when directly truncating teacher-generated reasoning trajectories (\method{} w/o Adaptive Read), the performance of \method{} decreases, indicating that reading the entire reasoning trajectories makes it difficult to localize a concise yet sufficient boundary. Finally, although \method{} w/o Binary Search achieves performance comparable to \method{}, it incurs substantially higher computational cost, resulting in more than 20 times higher latency than \method{} when searching for the truncation point (More details are provided in Appendix~\ref{app:effectiveness-binary-search}).


\subsection{Distillation Performance with Different Prefix Truncation Strategies\label{sec:5.3}}
\begin{figure}[t]
  \centering
\subfigure[AIME24\&25.\label{fig:effective-of-price:aime}]
{\includegraphics[width=0.23\textwidth]{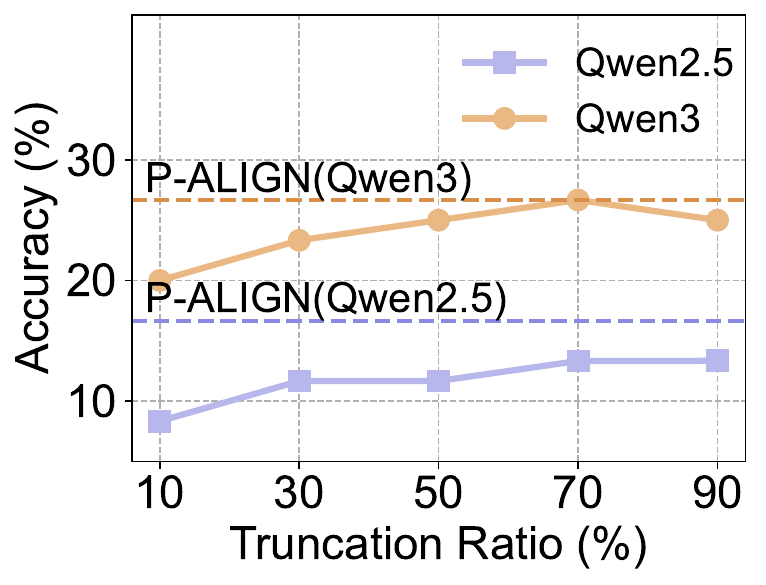}}
\subfigure[MATH500.\label{fig:effective-of-price:math500}]
{\includegraphics[width=0.23\textwidth]{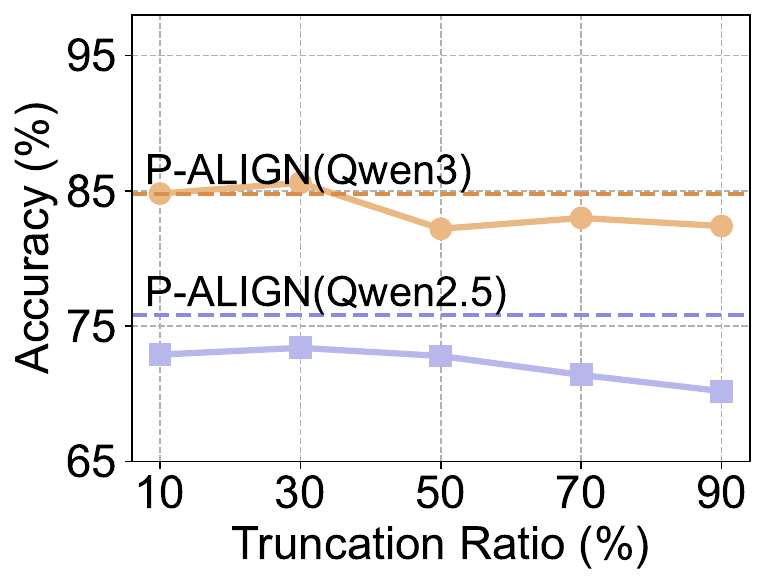}}
\subfigure[Response Length.\label{fig:effective-of-price:response-length}]
{\includegraphics[width=0.23\textwidth]{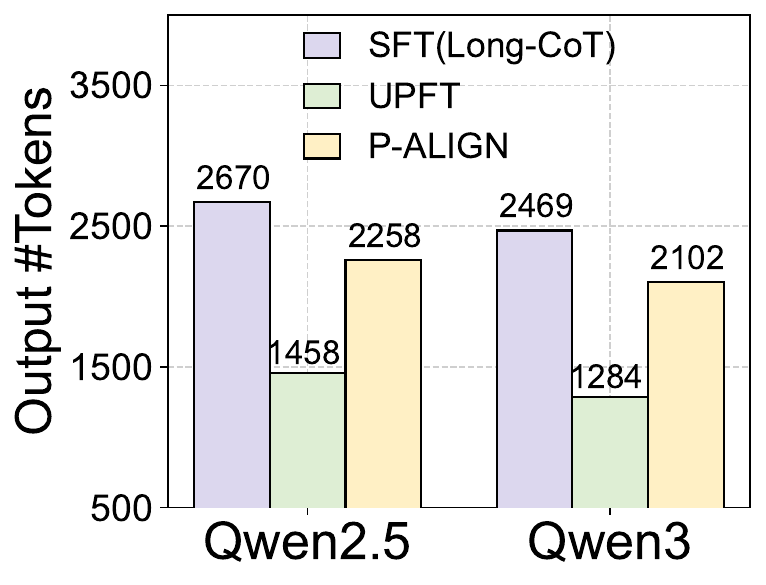}}
\subfigure[GLM Preference Rate.\label{fig:effective-of-price:gpt_score}]
{\includegraphics[width=0.23\textwidth]{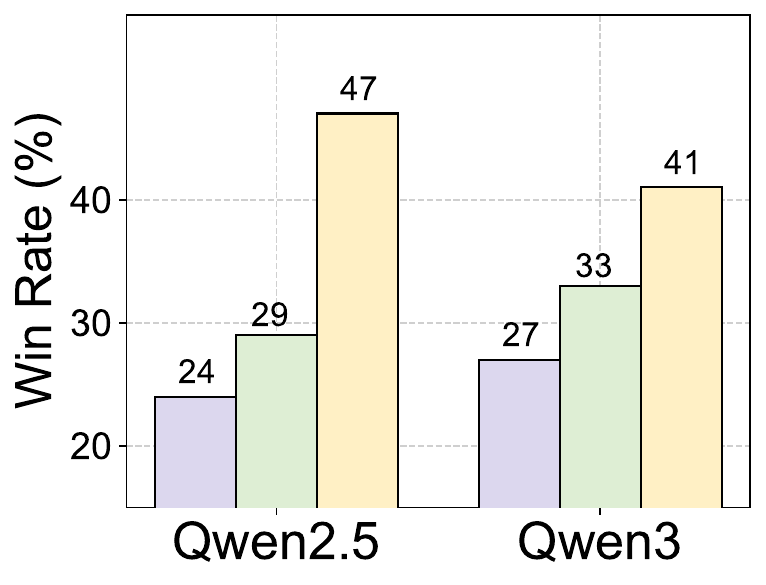}}

\caption{Performance of Distilled Student Models under Different Prefix Truncation Strategies. We compare models distilled using fixed prefix truncation ratios and \method{} (Figures~\ref{fig:effective-of-price:aime} and \ref{fig:effective-of-price:math500}), analyze their response lengths (Figure~\ref{fig:effective-of-price:response-length}), and evaluate CoT quality with GLM-4.5 as the judge (Figure~\ref{fig:effective-of-price:gpt_score}).}\label{fig:evaluate-price}
\end{figure}

In this section, we investigate the distillation performance under different prefix truncation strategies. As shown in Figure~\ref{fig:evaluate-price}, we first compare fixed-ratio truncation to motivate the necessity of adaptive prefix selection, and then analyze the length and quality of responses generated by models optimized using different prefix truncation strategies.

\textbf{Optimization with Fixed Prefix Ratios.} To evaluate the effectiveness of our adaptive prefix truncation method in \method{}, we compare \method{} with fixed ratio-based truncation strategies, where the truncation ratio ranges from 10\% to 90\%.

As shown in Figure~\ref{fig:effective-of-price:aime}, on competition-level benchmarks (AIME24\&25), the performance of optimized LLMs generally improves as longer prefixes are retained. In contrast, on the relatively easier dataset MATH500, performance degrades when longer prefixes are used (Figure~\ref{fig:effective-of-price:math500}). This suggests that, for difficult problems, longer reasoning trajectories provide informative and effective supervision signals for student model learning. Conversely, for easier problems, extended reasoning trajectories may introduce the overthinking issue~\cite{chen2024not}, which can mislead the student model and degrade its reasoning capability~\cite{luo2025through}. Benefiting from adaptive prefix truncation, \method{} consistently outperforms models trained with fixed-ratio truncation strategies across different settings, highlighting the critical role of adaptive truncation in balancing insufficient and excessive prefixes to deliver more effective supervision signals.

\textbf{Distillated Models with Different Truncation Strategies.}
To more thoroughly evaluate the effectiveness of optimized student models under different truncation strategies, we further analyze both the response lengths and the quality of the generated reasoning trajectories produced by these optimized student models.

We first present the average lengths of reasoning trajectories in Figure~\ref{fig:effective-of-price:response-length}. Among all models, UPFT produces significantly shorter trajectories, indicating that naively using prefixes as supervision signals may cause the student model to overfit superficial patterns. Compared with SFT (Long-CoT), \method{} generates more concise reasoning trajectories, demonstrating its effectiveness in mitigating the influence of redundant or uncertain reasoning content. We then employ GLM-4.5 (Figure~\ref{fig:effective-of-price:gpt_score}) to evaluate the quality of reasoning trajectories generated by different models, with detailed evaluation prompts provided in Appendix~\ref{app:prompt_evaluate_cots}. The evaluation results show that \method{} achieves the highest win rate, further confirming its effectiveness in providing higher-quality supervision that enables the student model to produce more concise and higher-quality reasoning trajectories.

\begin{figure}[t]
  \centering
\subfigure[Chunk Uncertainty at Different Positions.\label{fig:effective-of-prefix:entropy}]
{\includegraphics[width=0.23\textwidth]{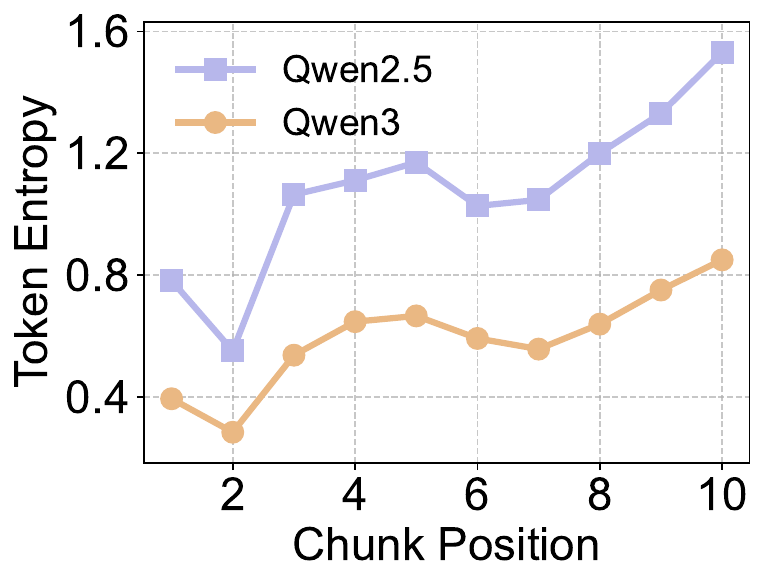}}
\subfigure[\method{} Using Fixed Teacher-generated Prefix.\label{fig:effective-of-prefix:Positional}]
{\includegraphics[width=0.23\textwidth]{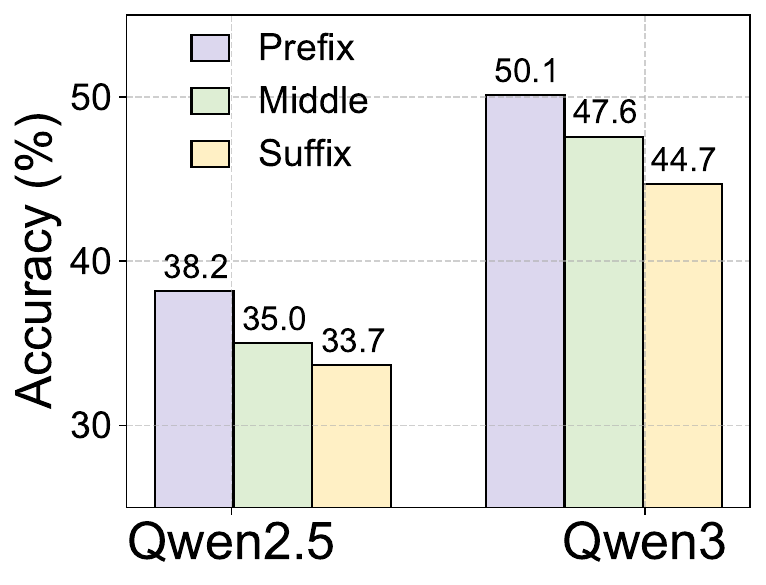}}
\caption{Effectiveness of Chunks in Long-Form CoTs at Different Positions. Figure~\ref{fig:effective-of-prefix:entropy} shows the entropy scores of student models across different chunks at varying positions. Figure~\ref{fig:effective-of-prefix:Positional} compares the performance of student models when using prefix, middle, or suffix segments as supervision.}\label{fig:evaluate-prefix}
\end{figure}

\subsection{Effectiveness of Reasoning Trajectory Chunks at Different Positions}
As illustrated in Figure~\ref{fig:evaluate-prefix}, we analyze the effectiveness of teacher-generated reasoning trajectories in supervising student models, with a particular focus on evaluating the contributions of trajectory prefixes at different positions.

As shown in Figure~\ref{fig:effective-of-prefix:entropy}, we assess the uncertainty of teacher-generated reasoning regions at different positions using the student model. Specifically, we evenly divide the teacher-generated reasoning trajectories into ten sequential chunks and compute the average token-level entropy score for chunks at each position. A higher entropy score indicates greater uncertainty~\cite{kendall2017uncertainties}. The results show that entropy increases as the chunk position moves toward the rear of the trajectory, indicating that later reasoning steps contain more uncertain content and potential exploratory trails. This observation is consistent with previous work~\cite{UPFT}.
Furthermore, as shown in Figure~\ref{fig:effective-of-prefix:Positional}, we evenly partition the teacher-generated CoT into three segments: prefix, middle, and suffix, and directly use each segment to replace the adaptively truncated prefix in \method{} for optimizing the student model. The results demonstrate that conditioning on prefixes yields substantially better performance than using either the middle or suffix segments, suggesting that prefixes provide more stable and informative context than later parts of the reasoning trajectory. This finding further validates the motivation of \method{} to fully exploit prefixes for synthesizing higher-quality supervision, thereby more effectively guiding the student model through SFT.
\section{Conclusion}

This paper proposes the Adaptive \textbf{P}refix-\textbf{ALIGN}ment reasoning distillation method (\method), which adaptively truncates prefixes from teacher-generated reasoning trajectories and optimizes student models to align with the retained teacher prefixes. 
Our experimental results demonstrate that \method{} consistently outperforms all baselines across multiple mathematical reasoning benchmarks, highlighting its effectiveness and robustness in distillating long-form reasoning into student models.

\section*{Limitation}
Although \method{} effectively constructs higher-quality SFT data, it still relies on powerful closed-source reasoning models to generate long-form reasoning chains, which incurs substantial computational overhead. Moreover, while adaptive prefix truncation proves effective, the self-judge process heavily depends on the judgment capability of the student model. This dependency may cause smaller-scale student models to become confused during self-judgment. Therefore, beyond self-information requirements, explicitly accounting for prefix quality is also crucial for adaptive prefix truncation.

\bibliography{arr2026-output}

@inproceedings{brown2020languagemodelsfewshotlearners,
 author = {Tom B. Brown and
Benjamin Mann and
Nick Ryder and
Melanie Subbiah and
Jared Kaplan and
Prafulla Dhariwal and
Arvind Neelakantan and
Pranav Shyam and
Girish Sastry and
Amanda Askell and
Sandhini Agarwal and
Ariel Herbert{-}Voss and
Gretchen Krueger and
Tom Henighan and
Rewon Child and
Aditya Ramesh and
Daniel M. Ziegler and
Jeffrey Wu and
Clemens Winter and
Christopher Hesse and
Mark Chen and
Eric Sigler and
Mateusz Litwin and
Scott Gray and
Benjamin Chess and
Jack Clark and
Christopher Berner and
Sam McCandlish and
Alec Radford and
Ilya Sutskever and
Dario Amodei},
 booktitle = {Proceedings of NeurIPS},
 title = {Language Models are Few-Shot Learners},
 url = {https://proceedings.neurips.cc/paper/2020/hash/1457c0d6bfcb4967418bfb8ac142f64a-Abstract.html},
 year = {2020}
}

@article{zhang2022optopenpretrainedtransformer,
 author = {Zhang, Susan and Roller, Stephen and Goyal, Naman and Artetxe, Mikel and Chen, Moya and Chen, Shuohui and Dewan, Christopher and Diab, Mona and Li, Xian and Lin, Xi Victoria},
 journal = {ArXiv preprint},
 title = {OPT: Open Pre-trained Transformer Language Models},
 url = {https://arxiv.org/abs/2205.01068},
 year = {2022}
}

@article{cobbe2021trainingverifierssolvemath,
 author = {Karl Cobbe and Vineet Kosaraju and Mohammad Bavarian and Mark Chen and Heewoo Jun and Lukasz Kaiser and Matthias Plappert and Jerry Tworek and Jacob Hilton and Reiichiro Nakano and Christopher Hesse and John Schulman},
 journal = {ArXiv preprint},
 title = {Training Verifiers to Solve Math Word Problems},
 url = {https://arxiv.org/abs/2110.14168},
 year = {2021}
}

@article{hendrycks2021measuringmathematicalproblemsolving,
 author = {Dan Hendrycks and Collin Burns and Saurav Kadavath and Akul Arora and Steven Basart and Eric Tang and Dawn Song and Jacob Steinhardt},
 journal = {ArXiv preprint},
 title = {Measuring Mathematical Problem Solving With the MATH Dataset},
 url = {https://arxiv.org/abs/2103.03874},
 year = {2021}
}

@inproceedings{zhang2024wrong,
 author = {Zhang, Yongheng and Chen, Qiguang and Zhou, Jingxuan and Wang, Peng and Si, Jiasheng and Wang, Jin and Lu, Wenpeng and Qin, Libo},
 booktitle = {Proceedings of EMNLP Findings},
 pages = {6644--6653},
 title = {Wrong-of-Thought: An Integrated Reasoning Framework with Multi-Perspective Verification and Wrong Information},
 url = {https://aclanthology.org/2024.findings-emnlp.388.pdf},
 year = {2024}
}

@inproceedings{hulora,
 author = {Edward J. Hu and
Yelong Shen and
Phillip Wallis and
Zeyuan Allen{-}Zhu and
Yuanzhi Li and
Shean Wang and
Lu Wang and
Weizhu Chen},
 booktitle = {Proceedings of ICLR},
 title = {LoRA: Low-Rank Adaptation of Large Language Models},
 url = {https://openreview.net/forum?id=nZeVKeeFYf9},
 year = {2022}
}

@article{wang2024learning,
 author = {Wang, Renxi and Li, Haonan and Han, Xudong and Zhang, Yixuan and Baldwin, Timothy},
 journal = {ArXiv preprint},
 title = {Learning From Failure: Integrating Negative Examples when Fine-tuning Large Language Models as Agents},
 url = {https://arxiv.org/abs/2402.11651},
 year = {2024}
}

@inproceedings{hsieh2023distillingstepbystepoutperforminglarger,
 author = {Hsieh, Cheng-Yu  and
Li, Chun-Liang  and
Yeh, Chih-kuan  and
Nakhost, Hootan  and
Fujii, Yasuhisa  and
Ratner, Alex  and
Krishna, Ranjay  and
Lee, Chen-Yu  and
Pfister, Tomas},
 booktitle = {Proceedings of ACL Findings},
 pages = {8003--8017},
 title = {Distilling Step-by-Step! Outperforming Larger Language Models with Less Training Data and Smaller Model Sizes},
 url = {https://aclanthology.org/2023.findings-acl.507},
 year = {2023}
}

@article{reich2023overcome,
 author = {Reich, Taly and Kaju, Alex and Maglio, Sam J},
 journal = {ArXiv preprint},
 title = {How to overcome algorithm aversion: Learning from mistakes},
 url = {https://arxiv.org/abs/10.1002},
 year = {2010}
}

@article{luo2023empirical,
  title={An empirical study of catastrophic forgetting in large language models during continual fine-tuning},
  author={Luo, Yun and Yang, Zhen and Meng, Fandong and Li, Yafu and Zhou, Jie and Zhang, Yue},
  journal={IEEE Transactions on Audio, Speech and Language Processing},
  url = {https://ieeexplore.ieee.org/document/11151751},
  year={2025},
  publisher={IEEE}
}

@article{gudibande2023false,
 author = {Gudibande, Arnav and Wallace, Eric and Snell, Charlie and Geng, Xinyang and Liu, Hao and Abbeel, Pieter and Levine, Sergey and Song, Dawn},
 journal = {ArXiv preprint},
 title = {The false promise of imitating proprietary llms},
 url = {https://arxiv.org/abs/2305.15717},
 year = {2023}
}

@article{wettig2024qurating,
 author = {Wettig, Alexander and Gupta, Aatmik and Malik, Saumya and Chen, Danqi},
 journal = {ArXiv preprint},
 title = {Qurating: Selecting high-quality data for training language models},
 url = {https://arxiv.org/abs/2402.09739},
 year = {2024}
}

@article{wang2023making,
 author = {Wang, Peiyi and Li, Lei and Chen, Liang and Song, Feifan and Lin, Binghuai and Cao, Yunbo and Liu, Tianyu and Sui, Zhifang},
 journal = {ArXiv preprint},
 title = {Making large language models better reasoners with alignment},
 url = {https://arxiv.org/abs/2309.02144},
 year = {2023}
}

@article{li2023query,
 author = {Li, Chengpeng and Yuan, Zheng and Dong, Guanting and Lu, Keming and Wu, Jiancan and Tan, Chuanqi and Wang, Xiang and Zhou, Chang},
 journal = {ArXiv preprint},
 title = {Query and response augmentation cannot help out-of-domain math reasoning generalization},
 url = {https://arxiv.org/abs/2310.05506},
 year = {2023}
}

@inproceedings{wei2022chain,
 author = {Jason Wei and
Xuezhi Wang and
Dale Schuurmans and
Maarten Bosma and
Brian Ichter and
Fei Xia and
Ed H. Chi and
Quoc V. Le and
Denny Zhou},
 booktitle = {Proceedings of NeurIPS},
 title = {Chain-of-Thought Prompting Elicits Reasoning in Large Language Models},
 url = {http://papers.nips.cc/paper\_files/paper/2022/hash/9d5609613524ecf4f15af0f7b31abca4-Abstract-Conference.html},
 year = {2022}
}

@inproceedings{qin2023cross,
 author = {Qin, Libo  and
Chen, Qiguang  and
Wei, Fuxuan  and
Huang, Shijue  and
Che, Wanxiang},
 booktitle = {Proceedings of EMNLP},
 pages = {2695--2709},
 title = {Cross-lingual Prompting: Improving Zero-shot Chain-of-Thought Reasoning across Languages},
 url = {https://aclanthology.org/2023.emnlp-main.163},
 year = {2023}
}

@article{luo2023wizardmath,
 author = {Luo, Haipeng and Sun, Qingfeng and Xu, Can and Zhao, Pu and Lou, Jianguang and Tao, Chongyang and Geng, Xiubo and Lin, Qingwei and Chen, Shifeng and Zhang, Dongmei},
 journal = {ArXiv preprint},
 title = {Wizardmath: Empowering mathematical reasoning for large language models via reinforced evol-instruct},
 url = {https://arxiv.org/abs/2308.09583},
 year = {2023}
}

@article{qwen2025qwen25technicalreport,
 author = {An Yang and Baosong Yang and Beichen Zhang and Binyuan Hui.},
 journal = {ArXiv preprint},
 title = {Qwen2.5 Technical Report},
 url = {https://arxiv.org/abs/2412.15115},
 year = {2024}
}

@article{grattafiori2024llama3herdmodels,
 author = {Aaron Grattafiori and Abhimanyu Dubey and Abhinav Jauhri and Abhinav Pandey and Abhishek Kadian},
 journal = {ArXiv preprint},
 title = {The Llama 3 Herd of Models},
 url = {https://arxiv.org/abs/2407.21783},
 year = {2024}
}

@article{deepseekai2025deepseekr1incentivizingreasoningcapability,
 author = {DeepSeek-AI and Daya Guo and Dejian Yang and Haowei Zhang},
 journal = {ArXiv preprint},
 title = {DeepSeek-R1: Incentivizing Reasoning Capability in LLMs via Reinforcement Learning},
 url = {https://arxiv.org/abs/2501.12948},
 year = {2025}
}

@article{godahewa2021monashtimeseriesforecasting,
 author = {Rakshitha Godahewa and Christoph Bergmeir and Geoffrey I. Webb and Rob J. Hyndman and Pablo Montero-Manso},
 journal = {ArXiv preprint},
 title = {Monash Time Series Forecasting Archive},
 url = {https://arxiv.org/abs/2105.06643},
 year = {2021}
}

@article{yin2025marcoo1v2wideningdistillation,
 author = {Huifeng Yin and Yu Zhao and Minghao Wu and Xuanfan Ni and Bo Zeng and Hao Wang and Tianqi Shi and Liangying Shao and Chenyang Lyu and Longyue Wang and Weihua Luo and Kaifu Zhang},
 journal = {ArXiv preprint},
 title = {Marco-o1 v2: Towards Widening The Distillation Bottleneck for Reasoning Models},
 url = {https://arxiv.org/abs/2503.01461},
 year = {2025}
}

@article{li2025smallmodelsstrugglelearn,
 author = {Yuetai Li and Xiang Yue and Zhangchen Xu and Fengqing Jiang and Luyao Niu and Bill Yuchen Lin and Bhaskar Ramasubramanian and Radha Poovendran},
 journal = {ArXiv preprint},
 title = {Small Models Struggle to Learn from Strong Reasoners},
 url = {https://arxiv.org/abs/2502.12143},
 year = {2025}
}

@inproceedings{muennighoff2025s1simpletesttimescaling,
  title={s1: Simple test-time scaling},
  author={Muennighoff, Niklas and Yang, Zitong and Shi, Weijia and Li, Xiang Lisa and Fei-Fei, Li and Hajishirzi, Hannaneh and Zettlemoyer, Luke and Liang, Percy and Cand{\`e}s, Emmanuel and Hashimoto, Tatsunori B},
  booktitle={Proceedings of EMNLP},
  url = {https://aclanthology.org/2025.emnlp-main.1025/},
  pages={20286--20332},
  year={2025}
}

@article{ye2025limoreasoning,
 author = {Yixin Ye and Zhen Huang and Yang Xiao and Ethan Chern and Shijie Xia and Pengfei Liu},
 journal = {ArXiv preprint},
 title = {LIMO: Less is More for Reasoning},
 url = {https://arxiv.org/abs/2502.03387},
 year = {2025}
}

@article{yang2025qwen3technicalreport,
 author = {An Yang and Anfeng Li and Baosong Yang and Beichen Zhang, et al.},
 journal = {ArXiv preprint},
 title = {Qwen3 Technical Report},
 url = {https://arxiv.org/abs/2505.09388},
 year = {2025}
}

@inproceedings{anonymous2025morale,
 author = {Anonymous},
 booktitle = {Submitted to ACL Rolling Review - May 2025},
 note = {under review},
 title = {{MORALE}: Segment-Guided Distillation Framework for Small Reasoning Models},
 url = {https://openreview.net/forum?id=CWUpE9cQ6u},
 year = {2025}
}

@article{chen2024not,
 author = {Chen, Xingyu and Xu, Jiahao and Liang, Tian and He, Zhiwei and Pang, Jianhui and Yu, Dian and Song, Linfeng and Liu, Qiuzhi and Zhou, Mengfei and Zhang, Zhuosheng and others},
 journal = {ArXiv preprint},
 title = {Do not think that much for 2+ 3=? on the overthinking of o1-like llms},
 url = {https://arxiv.org/abs/2412.21187},
 year = {2024}
}

@article{AMC12,
 howpublished = {\url{https://artofproblemsolving.com/wiki/index.php/AMC_12_Problems_and_Solutions}},
 url = {https://artofproblemsolving.com/wiki/index.php/AMC_12_Problems_and_Solutions},
 note = {Accessed: September 23, 2025},
 title = {{AMC12}. AMC 12 Problems and Solutions},
 year = {2025}
}

@article{AIME,
 howpublished = {\url{https://artofproblemsolving.com/wiki/index.php/AIME_Problems_and_Solutions}},
url = {https://artofproblemsolving.com/wiki/index.php/AIME_Problems_and_Solutions},
 note = {Accessed: September 23, 2025},
 title = {{AIME}. AIME Problems and Solutions},
 year = {2025}
}

@article{UPFT,
 author = {Ji, Ke and {Jiahao Xu} and Liang, Tian and {Qiuzhi Liu} and {Zhiwei He} and {Xingyu Chen} and {Xiaoyuan Liu} and {Zhijie Wang} and {Junying Chen} and {Benyou Wang} and {Zhaopeng Tu} and {Haitao Mi} and Yu, Dong},
 language = {en},
 title = {The First Few Tokens Are All You Need: An Efficient and Effective Unsupervised Prefix Fine-Tuning Method for Reasoning Models},
 url = {https://www.researchgate.net/doi/10.13140/RG.2.2.33772.07043},
 year = {2025}
}

@article{chen2025datacentricsamplecentricenhancingllm,
 author = {Xinjie Chen and Minpeng Liao and Guoxin Chen and Chengxi Li and Biao Fu and Kai Fan and Xinggao Liu},
 journal = {ArXiv preprint},
 title = {From Data-Centric to Sample-Centric: Enhancing LLM Reasoning via Progressive Optimization},
 url = {https://arxiv.org/abs/2507.06573},
 year = {2025}
}

@article{chen2025towards,
 author = {Chen, Qiguang and Qin, Libo and Liu, Jinhao and Peng, Dengyun and Guan, Jiannan and Wang, Peng and Hu, Mengkang and Zhou, Yuhang and Gao, Te and Che, Wanxiang},
 journal = {ArXiv preprint},
 title = {Towards reasoning era: A survey of long chain-of-thought for reasoning large language models},
 url = {https://arxiv.org/abs/2503.09567},
 year = {2025}
}

@inproceedings{luo2025through,
 author = {Luo, Renjie and Li, Jiaxi and Huang, Chen and Lu, Wei},
 booktitle = {Proceedings of EMNLP},
 pages = {4972--4992},
 title = {Through the Valley: Path to Effective Long {C}o{T} Training for Small Language Models},
 url = {https://aclanthology.org/2025.emnlp-main.251/},
 year = {2025}
}

@article{xu2025chain,
 author = {Xu, Silei and Xie, Wenhao and Zhao, Lingxiao and He, Pengcheng},
 journal = {ArXiv preprint},
 title = {Chain of draft: Thinking faster by writing less},
 url = {https://arxiv.org/abs/2502.18600},
 year = {2025}
}

@article{wu2025enhancing,
 author = {Wu, Zhuoyang and Li, Xinze and Liu, Zhenghao and Yan, Yukun and Liu, Zhiyuan and Yu, Minghe and Yang, Cheng and Gu, Yu and Yu, Ge and Sun, Maosong},
 journal = {ArXiv preprint},
 title = {Enhancing Long-Chain Reasoning Distillation through Error-Aware Self-Reflection},
 url = {https://arxiv.org/abs/2505.22131},
 year = {2025}
}

@inproceedings{wang2025infogain,
 author = {Wang, Zihan and Liang, Zihan and Shao, Zhou and Ma, Yufei and Dai, Huangyu and Chen, Ben and Mao, Lingtao and Lei, Chenyi and Ding, Yuqing and Li, Han},
 booktitle = {Proceedings of EMNLP},
 title = {InfoGain-RAG: Boosting Retrieval-Augmented Generation through Document Information Gain-based Reranking and Filtering},
 year = {2025}
}

@article{zeng2025pruning,
 author = {Zeng, Wenhao and Wang, Yaoning and Hu, Chao and Shi, Yuling and Wan, Chengcheng and Zhang, Hongyu and Gu, Xiaodong},
 journal = {ArXiv preprint},
 title = {Pruning the unsurprising: Efficient code reasoning via first-token surprisal},
 url = {https://arxiv.org/abs/2508.05988},
 year = {2025}
}

@article{jin2025recut,
 author = {Jin, Zhensheng and Li, Xinze and Ji, Yifan and Peng, Chunyi and Liu, Zhenghao and Shi, Qi and Yan, Yukun and Wang, Shuo and Peng, Furong and Yu, Ge},
 journal = {ArXiv preprint},
 title = {ReCUT: Balancing Reasoning Length and Accuracy in LLMs via Stepwise Trails and Preference Optimization},
 url = {https://arxiv.org/abs/2506.10822},
 year = {2025}
}

@article{zeng2025glm,
 author = {Zeng, Aohan and Lv, Xin and Zheng, Qinkai and Hou, Zhenyu and Chen, Bin and Xie, Chengxing and Wang, Cunxiang and Yin, Da and Zeng, Hao and Zhang, Jiajie and others},
 journal = {ArXiv preprint},
 title = {Glm-4.5: Agentic, reasoning, and coding (arc) foundation models},
 url = {https://arxiv.org/abs/2508.06471},
 year = {2025}
}

@inproceedings{kendall2017uncertainties,
 author = {Alex Kendall and
Yarin Gal},
 booktitle = {Proceedings of NeurIPS},
 pages = {5574--5584},
 title = {What Uncertainties Do We Need in Bayesian Deep Learning for Computer
Vision?},
 url = {https://proceedings.neurips.cc/paper/2017/hash/2650d6089a6d640c5e85b2b88265dc2b-Abstract.html},
 year = {2017}
}

@article{sun2025begunhalfdonereinforcement,
 author = {Yiliu Sun and Zicheng Zhao and Yang Wei and Yanfang Zhang and Chen Gong},
 journal = {ArXiv preprint},
 title = {Well Begun, Half Done: Reinforcement Learning with Prefix Optimization for LLM Reasoning},
 url = {https://arxiv.org/abs/2512.15274},
 year = {2025}
}

@article{zelikman2022star,
 author = {Zelikman, Eric and Wu, Yuhuai and Mu, Jesse and Goodman, Noah},
 journal = {Advances in Neural Information Processing Systems},
 pages = {15476--15488},
 title = {Star: Bootstrapping reasoning with reasoning},
 year = {2022}
}
\clearpage

\appendix
\section{Appendix}
\subsection{License}
We provide the licenses for the datasets used in \method{}. s1K-1.1 is licensed under the MIT License, while MATH500, AIME24, AIME25, and AMC12 are licensed under the Apache License 2.0. All of these licenses permit the use of their data for academic purposes.

\begin{table}[t]
\centering
\small
\begin{tabular}{l|r r}
\hline
\textbf{Dataset} & \textbf{Train}  & \textbf{Test} \\  
\hline
s1K-1.1~\shortcite{muennighoff2025s1simpletesttimescaling} & 1,000  & - \\
AIME25~\shortcite{AIME} & -  & 30 \\        AIME24~\shortcite{AIME} & - & 30 \\  
AMC23~\shortcite{AMC12} & - & 83 \\ 
MATH500~\shortcite{godahewa2021monashtimeseriesforecasting} & - & 500 \\ 

\hline
\end{tabular}
\caption{Data Statistics.}
\label{tab:dataset}
\end{table}
\begin{table}[t]
\centering
\small
\renewcommand{\arraystretch}{1.2} 

\begin{tabular}{l|c|c|c}

\hline
\textbf{Model} & \textbf{Long-CoT} & \textbf{Prefix} & \textbf{\method{}}  \\
\hline
Qwen2.5-7B  & 9,291 & 4,531 & 5,453 \\
Qwen3-8B    & 9,291 & 4,402 & 5,732 \\
Llama3.2-3B & 8,755 & 4,780 & 6,035 \\
\hline
\end{tabular}
\caption{Average Token Length Statistics of Training Dataset Across Different Student Models. We evaluate the average token lengths of the original long-form CoTs, the truncated prefixes, and the final trajectories used by \method{}.}
\label{tab:data-static-token}
\end{table}

\subsection{Additional Implementation Details}\label{app:data-statistics}

In this section, we provide the detailed data statistics in \method{}. We build our training set using 1,000 examples sampled from s1K-1.1~\cite{muennighoff2025s1simpletesttimescaling} as the training dataset. For evaluation, we use the MATH500 with 500 examples, AIME25 with 30 examples, AIME24 with 30 examples, and AMC12 with 83 examples. All statistics are shown in Table~\ref{tab:dataset}.

To clarify the impact of \method{} on supervision efficiency, we report average token length statistics across different student models in Table~\ref{tab:data-static-token}. Specifically, Long-CoT denotes the token length of the original teacher-generated reasoning traces, Prefix corresponds to the adaptive prefixes truncated by \method{}, and \method{} reports the token length of the final trajectories used for SFT. The results show that \method{} substantially reduces the length of training data by retaining only concise and sufficient prefixes, which lowers training cost while maintaining or even improving model performance.

\subsection{Prompt Templates Used in \method{}}\label{app:prompt-price}

\begin{figure*}[t]
    \centering    \includegraphics[width=0.9\linewidth]{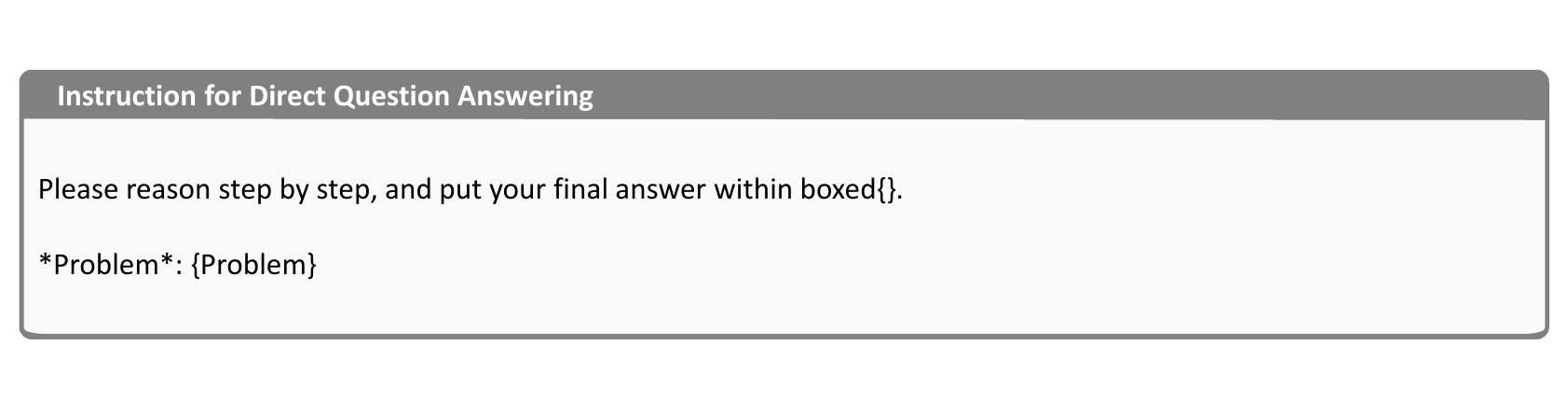}
    \caption{The Prompt Templates Used for Direct Question Answering.}
    \label{fig:prompt-math_qa}
\end{figure*}

\begin{figure*}[t]
    \centering    \includegraphics[width=0.9\linewidth]{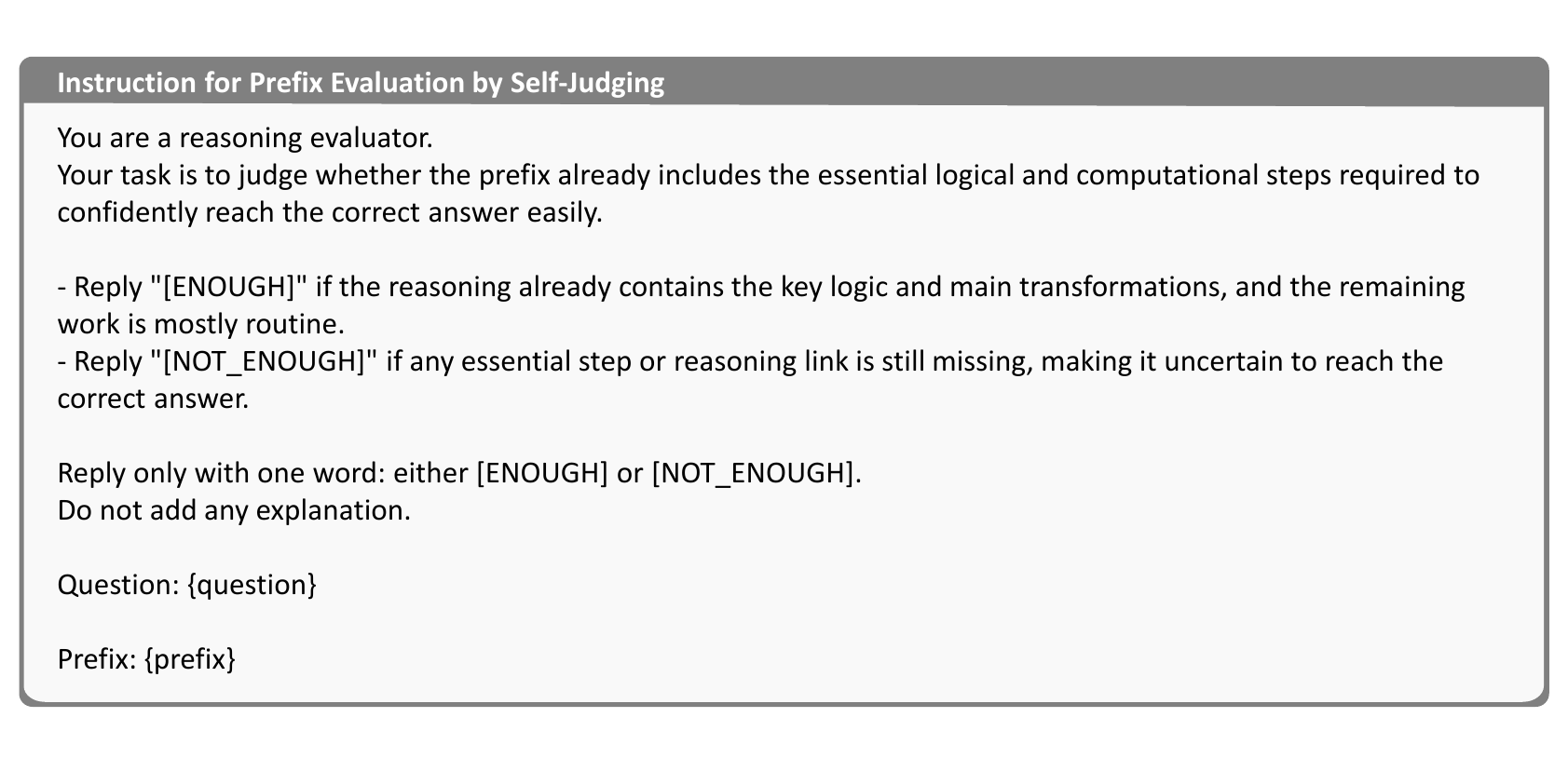}
    \caption{The Prompt Templates Used for Prefix Evaluation.}
    \label{fig:prompt-prefix_eval}
\end{figure*}

\begin{figure*}[t]
    \centering
    \includegraphics[width=0.9\linewidth]{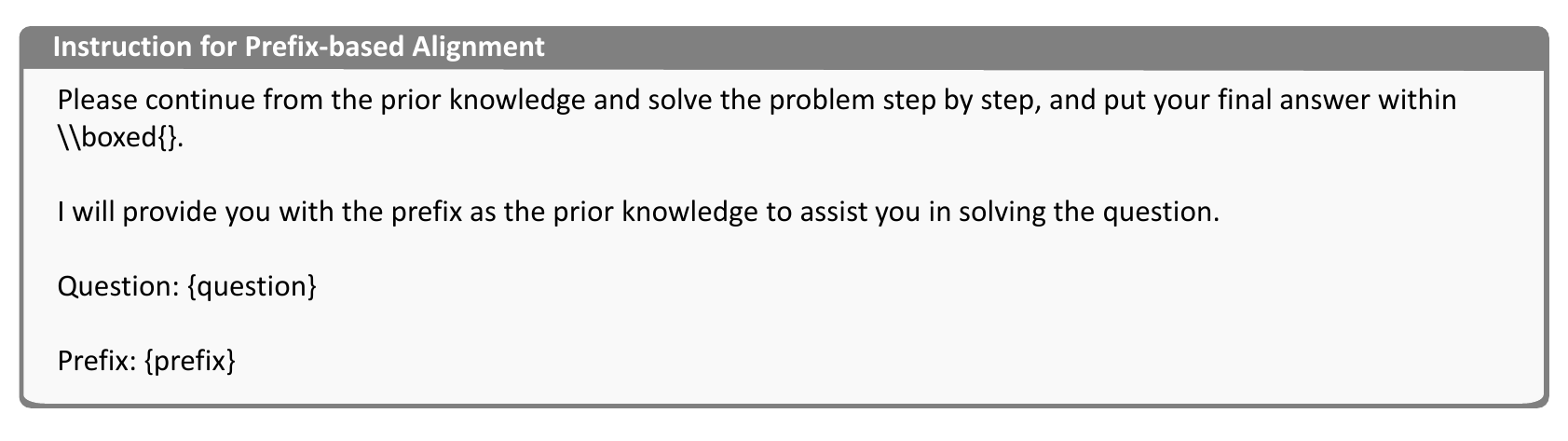}
    \caption{The Prompt Templates Used for Prefix-based Alignment.}
    \label{fig:prompt-cot_completion}
\end{figure*}

We detail the instruction prompts employed at different stages of \method{}. As shown in Figure~\ref{fig:prompt-math_qa}, the $\text{Instruct}_\text{QA}$ prompts the model to directly answer the given question, serving as the standard question-answering setting and providing the initial reasoning. Building on this, the prefix evaluation stage utilizes the $\text{Instruct}_\text{Eval}$ prompt, as illustrated in Figure~\ref{fig:prompt-prefix_eval}, which guides the student model in assessing whether a given reasoning prefix contains sufficient information to solve the problem. Based on the retained prefix, the prefix-based alignment stage applies the $\text{Instruct}_\text{Align}$ prompt shown in Figure~\ref{fig:prompt-cot_completion}, guiding the student model to continue generating a complete reasoning trajectory.

\subsection{Prompt Templates Used for Evaluating the Quality of CoTs }\label{app:prompt_evaluate_cots}

\begin{figure*}[t]
    \centering
    \includegraphics[width=0.9\linewidth]{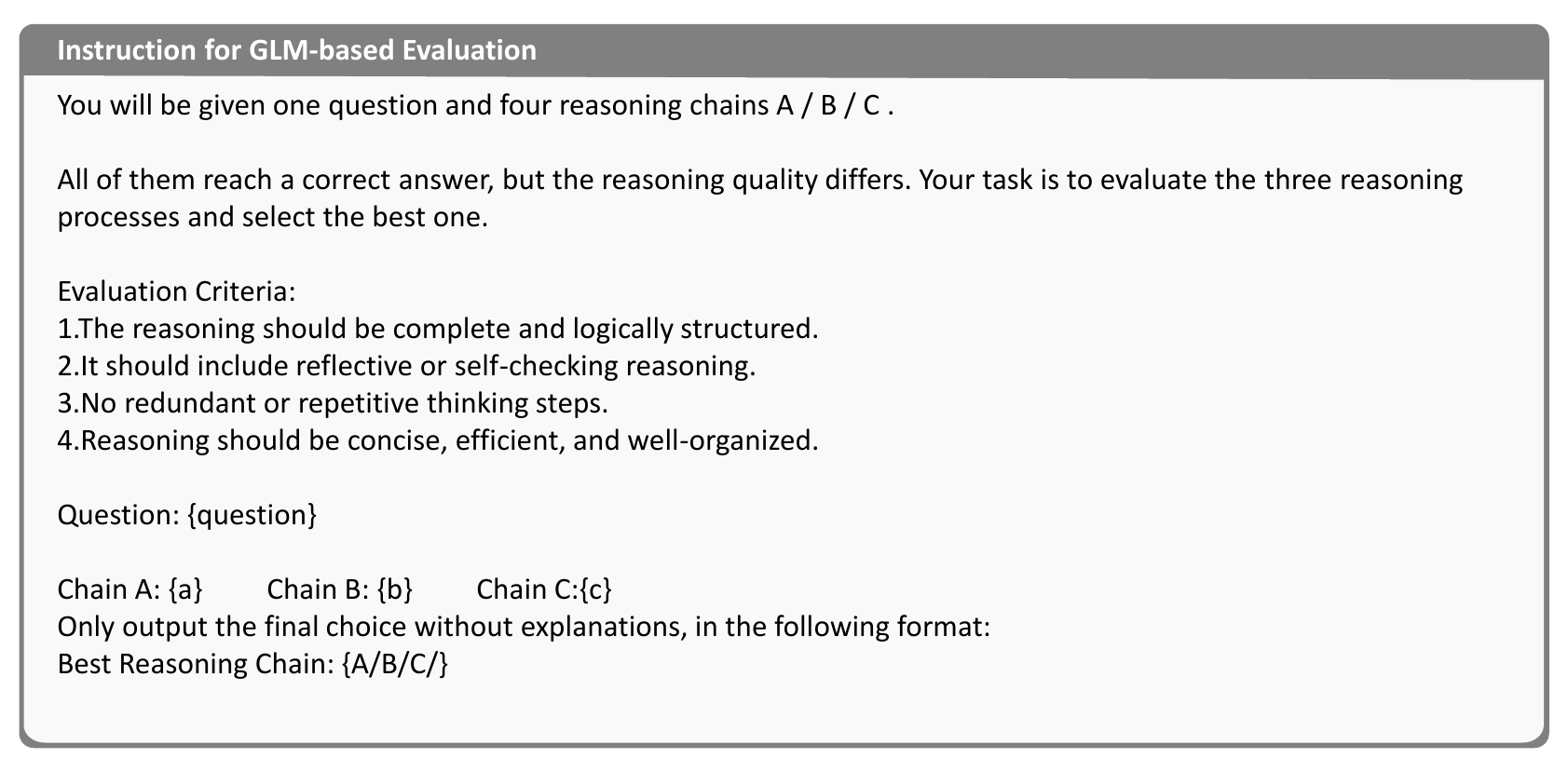}
    \caption{The Prompt Templates Used for GLM-based Evaluation.}
    \label{fig:prompt-glm_based_evaluation}
\end{figure*}

As shown in Figure~\ref{fig:prompt-glm_based_evaluation}, we present the prompt templates used to evaluate the quality of the CoTs synthesized by three methods: SFT (Long-CoT), UPFT, and \method{}. The evaluation is conducted using GLM-4.5~\cite{zeng2025glm} as the evaluator, which assesses each generated reasoning trace based on its logical completeness, reflective reasoning behavior, conciseness, and overall organization. We employ consistent prompt templates across all methods to ensure fair comparison and reproducibility.

\begin{table}[t]
\small
\centering
\setlength{\tabcolsep}{3.2pt} 
\renewcommand{\arraystretch}{1.2} 
\footnotesize 
\begin{tabular}{l|cccc}
\hline
\textbf{Methods}  &\textbf{AIME24}  &\textbf{AMC12} &\textbf{MATH500} &\textbf{Avg.}  \\ \hline
\rowcolor{gray!8}\multicolumn{5}{l}{\textbf{\textit{Llama3.2-3B}}} \\
Zero-Shot & 6.67 & 18.81 & 39.80 & 21.76 \\ 
SFT (Long-CoT) & 6.67 & 20.19 & 43.60 & 23.49 \\ 
UPFT~\shortcite{UPFT} &10.00 & 26.50 & 44.60 & 27.03 \\
\method{} & \textbf{16.67} & \textbf{30.12} & \textbf{48.40} & \textbf{31.73} \\ \hline
\rowcolor{gray!8}\multicolumn{5}{l}{\textbf{\textit{Qwen3-14B}}} \\
Zero-Shot & 33.33 & 72.29 & 87.40 & 64.34\\ 
SFT  & 36.67 & 73.26 & 88.60 & 66.18   \\ 
UPFT~\shortcite{UPFT} & 36.67 & \textbf{74.69} & \textbf{89.60} & 66.99\\
\method{}  & \textbf{43.33} & 73.26 & 89.20 & \textbf{68.60}   \\ \hline

\end{tabular}
    \caption{Performance of \method{} on Models of Different Sizes. We provide additional results on Llama3.2-3B and Qwen3-14B to evaluate the generalization of \method{} across different parameter scales.}
\label{tab:appendix-result-other-model}
\end{table}

\begin{table}[t]
\centering
\small
\renewcommand{\arraystretch}{1.2} 

\begin{tabular}{l|c|c}
\hline
\textbf{Backbone Model} & \textbf{\method{}} & \textbf{w/o Binary Search}  \\
\hline
Qwen2.5-7B  & 7.4  & 144.6 \\
Qwen3-8B    & 7.4  & 137.1 \\
Llama3.2-3B & 7.4  & 152.9 \\
\hline
\end{tabular}
\caption{Average Search Count for Binary Search Usage in Adaptive Prefix Truncation.}
\label{tab:average-number-search}
\end{table}

\subsection{Verification of Binary Search Effectiveness in \method{}}\label{app:effectiveness-binary-search}

To further verify the effectiveness of binary search in improving the efficiency of adaptive prefix truncation, we measure the average search count for determining whether binary search is used in prefix truncation. When binary search is not used, the sufficient condition is checked in a sentence-level sequential manner. As shown in Table~\ref{tab:average-number-search}, the average search count without binary search is about 20 times higher than when binary search is used. This clearly demonstrates the effectiveness of binary search in improving efficiency. Notably, the total number of iterations for binary search is fixed at \( O(\log_2 n) \), independent of both the student model and the results of the \texttt{ENOUGH} judgment.

\subsection{Additional Results with Different Student Models}\label{app:other_student_models}
To further evaluate the scalability and robustness of \method{}, we conduct additional experiments on models of different sizes, including the smaller Llama3.2-3B-Instruct~\cite{grattafiori2024llama3herdmodels} and the larger Qwen3-14B~\cite{yang2025qwen3technicalreport}. All training settings remain consistent with those used for the 7B-scale experiments. As shown in Table~\ref{tab:appendix-result-other-model}, \method{} yields consistent improvements across model scales, demonstrating that the proposed framework generalizes effectively to both lower-capacity and higher-capacity models.

\subsection{The Case of Binary Search for Minimal Sufficient Prefix Selection}
\label{app:binary_search_case}

\begin{figure*}[t]
    \centering
    \includegraphics[width=0.9\linewidth]{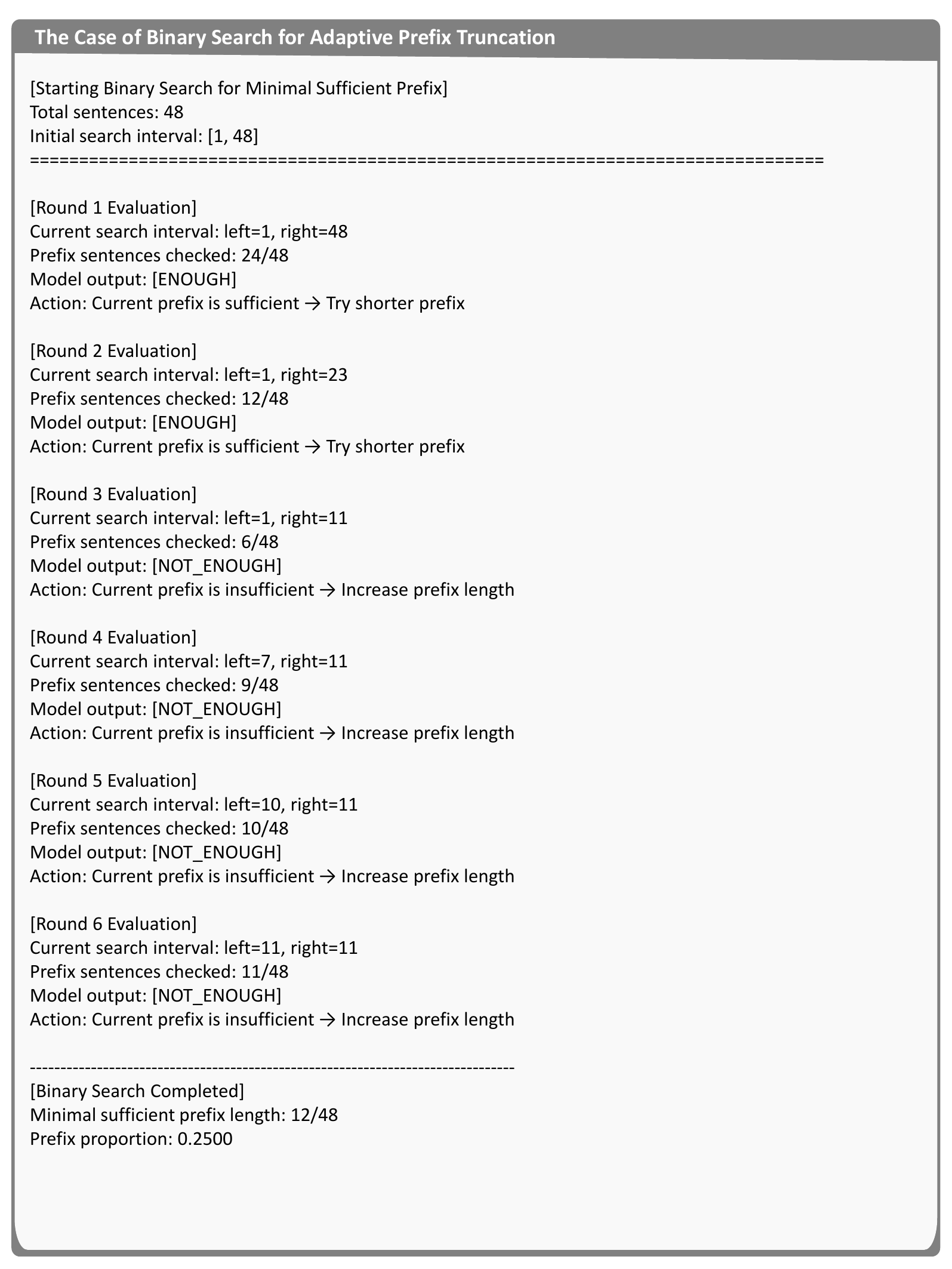}
    \caption{The Case of Adaptive Prefix Truncation via Binary Search.}
    \label{fig:binary-search-case}
\end{figure*}

As shown in Figure~\ref{fig:binary-search-case}, \method{} employs the binary search procedure for adaptive prefix evaluation by self-judging, which improves both the efficiency and precision of locating a minimal sufficient prefix. Specifically, when the self-judging evaluation returns \texttt{ENOUGH}, the current prefix already contains sufficient information to solve the problem, and we continue searching for a shorter prefix to obtain a more concise and learnable supervision signal. Otherwise, if the evaluation returns \texttt{NOT\_ENOUGH}, the current prefix is still insufficient, and a longer prefix with richer information is required to support solving the problem. In this case, \method{} identifies the minimal sufficient prefix with only $6$ self-evaluation iterations. In contrast, the sentence-by-sentence sequential scan (\method{} w/o Binary Search) requires $12$ evaluations to reach the same boundary. This comparison demonstrates that binary search substantially reduces the prefix-selection cost and avoids unnecessary evaluation overhead.

\subsection{Case Study}\label{app:case-study}
In this section, we present a detailed case study using two tables to illustrate the key steps and the qualitative benefits of \method{}. As shown in Table~\ref{case-study:method}, we illustrate the full pipeline of \method{}, including the prefix truncation and prefix-based alignment process. Building on this, Table~\ref{case-study:different_strategy} compares the final response generated by \method{} and baseline methods, highlighting the improved quality and clarity of \method{} outputs.

First, we examine a representative mathematical problem to analyze the behavior of \method{}. As shown in Table~\ref{case-study:method}, the key reasoning that directly supports the correct answer(``Therefore, $P \equiv 109 \pmod{125}$'') appears early in the teacher's reasoning trajectory, within the first 30\% of the full reasoning length. The subsequent steps often contain more reflective and exploratory reasoning, which can be less learnable for the student model. In this case, the vanilla student model initially fails and produces an incorrect result (``thus the remainder is 999''). In contrast, when conditioned on the retained prefix, the student can follow the preserved reasoning structure and complete the remaining reasoning steps, ultimately arriving at the correct solution (``Therefore, the remainder is boxed \{109\}.''). This comparison highlights the advantage of prefix-aligned supervision in stabilizing the student's reasoning and helping it recover correct solutions that are otherwise difficult to obtain.

Then, we compare student outputs under different baselines on the same question. As shown in Table~\ref{case-study:different_strategy}, SFT (Label) provides little explicit reasoning and directly outputs an incorrect answer. The vanilla model attempts step-by-step derivations, but fails to identify the key insight of applying Vieta's formula and instead attempts to solve the cubic equation explicitly, which is error-prone due to heavy algebraic computation.
Furthermore, SFT (Long-CoT) performs a deeper analysis and recognizes the crucial insight early (``That reminds me of Vieta's formula''), realizing that the problem can be solved without computing each root. However, excessive reflection in later steps still leads to a calculation error, despite having found the correct approach.
Compared with these baselines, \method{} distills the reasoning patterns through prefix-based supervision, enabling the student to follow the correct high-level structure without overthinking. As a result, it produces a cleaner solution trajectory and reaches the correct final answer (``Therefore, the new volume is boxed \{30\}.'').


\begin{table*}[t]
\centering
\small
\renewcommand\arraystretch{1.2}
\setlength{\fboxsep}{1.0pt}
\resizebox{\linewidth}{!}{ 
\begin{tabular}{p{0.88\textwidth}}
\hline
\rowcolor{gray!8}{\textbf{Case:} The Case of Prefix-Alignment Distillation in \method{}.}

\\
\hline
\textbf{Question:} Find the remainder when $9 \times 99 \times 999 \times \cdots \times \underbrace{99\cdots9}_{\text{999 9's}}$ is divided by $1000$.

\\
\textbf{Ground-Truth:} 109

\\
\textbf{Vanilla Generation:} 
To find the remainder of the question\ldots we can use properties of modular arithmetic and the Chinese Remainder Theorem\ldots Each number in the product is of the form \(10^k - 1\). For any \(n \geq 3\), \textbf{\textcolor[rgb]{0.7,0.3,0.3}{\(10^n \equiv 0 \pmod{8}\)}}, hence \textbf{\textcolor[rgb]{0.7,0.3,0.3}{\(10^n - 1 \equiv -1 \pmod{8}\)}}. Thus the product satisfies \textbf{\textcolor[rgb]{0.7,0.3,0.3}{\(9 \times 99 \times 999 \times \cdots \equiv (-1)^{999} \equiv -1 \pmod{8}\)}}\ldots Similarly, since each \(10^k - 1\) for \(k \geq 3\) ends in 999, the product modulo 125 is claimed to be \textbf{\textcolor[rgb]{0.7,0.3,0.3}{\(9 \times 99 \times 999 \times \cdots \equiv (-1)^{999} \equiv -1 \pmod{125}\)}}. \ldots and thus the remainder is \textbf{\textcolor[rgb]{0.7,0.3,0.3}{\(999\)}}.
\\

\textbf{Original Long-Form Reasoning:} 
Alright, so I have this problem here: I need to find the remainder when the product of numbers like $9, 99, 999$, and so on, up to a number with $999$ nines, is divided by 1000. Hmm, okay, let's\ldots  Therefore, the product is $\prod_{k=1}^{999} (10^k - 1)$\ldots Since $10 \equiv 2 \pmod{8}$, so $10^k \pmod{8}$ is $2^k \pmod{8}$\ldots All terms from $k=3$ onward are $-1 \pmod{8}$.Therefore, $P \equiv 5 \pmod{8}$\ldots Therefore, $(10^k - 1) \equiv -1 \pmod{125}$ for $k \ge 3$ \ldots \textbf{\textcolor{blue}{Therefore, $P \equiv 109 \pmod{125}$\ldots}} \\

First, the problem is asking for the remainder of a very large product divided by 1000. When dealing with remainders, especially modulo 1000, it often helps to look at the problem modulo 8 and modulo 125 separately, since 1000 is 8 * 125 and 8 and 125 are coprime. Then, using the Chinese Remainder Theorem, combine the results\ldots A number consisting of n 9s can be written as 10\^{}n - 1. \ldots\ The number with k 9s is 10\^{}k - 1. Therefore, the product is $\Pi_{k=1}^{999} (10^k - 1)$. \ldots\ We need to compute $P \bmod 1000$\ldots Compute $P \bmod 8$. Since 10 $\equiv$ 2 (mod 8), 10\^{}k (mod 8) is 2\^{}k (mod 8), so each term (10\^{}k - 1) (mod 8) is (2\^{}k - 1) (mod 8). \ldots\ 
For k=1: (2\^{}1 - 1) = 1 (mod 8). 
For k=2: (4 - 1) = 3 (mod 8).
For k $\ge$ 3: 2\^{}k $\equiv$ 0 (mod 8), so (2\^{}k - 1) $\equiv$ -1 (mod 8). 
\ldots\ Therefore, $P \equiv 5 \pmod{8}$\ldots

Alright, now onto modulo 125. \ldots\
10\^{}3 = 1000 $\equiv$ 0 (mod 125). Therefore, for all k $\ge$ 3, 10\^{}k $\equiv$ 0 (mod 125), so (10\^{}k - 1) $\equiv$ -1 (mod 125). \ldots\
Thus, $P \equiv 9 \cdot 99 \cdot (-1)^{997} \pmod{125}$. \ldots\ 
9*99 = 891, and 891 (mod 125) = 16, hence $P \equiv -16 \equiv 109 \pmod{125}$\ldots So summary so far: $P \equiv 5 \pmod{8}$ and $P \equiv 109 \pmod{125}$. Now need to find x such that $x \equiv 5 \pmod{8}$ and $x \equiv 109 \pmod{125}$. \ldots\
Let x = 125m + 109. Then 125m + 109 $\equiv$ 5 (mod 8). \ldots\ 
This gives 5m $\equiv$ 0 (mod 8), so m $\equiv$ 0 (mod 8). Therefore, x = 1000n + 109, hence $x \equiv 109 \pmod{1000}$\ldots

\ldots

\textbf{\textcolor{OliveGreen}{Therefore, the boxed answer is boxed\{109\}}}.
\\
\textbf{Truncation Prefix:} 
Alright, so I have this problem here: I need to find the remainder when the product of numbers like $9, 99, 999$, and so on, up to a number with $999$ nines, is divided by 1000. Hmm, okay, let's\ldots  Therefore, the product is $\prod_{k=1}^{999} (10^k - 1)$\ldots Since $10 \equiv 2 \pmod{8}$, so $10^k \pmod{8}$ is $2^k \pmod{8}$\ldots All terms from $k=3$ onward are $-1 \pmod{8}$.Therefore, $P \equiv 5 \pmod{8}$\ldots Therefore, $(10^k - 1) \equiv -1 \pmod{125}$ for $k \ge 3$ \ldots \textbf{\textcolor{blue}{Therefore, $P \equiv 109 \pmod{125}$}}.

\\
\textbf{Prefix-based Alignment:} 
To continue answering the question based on existing prior knowledge \ldots First, consider the product modulo $8$\ldots the product modulo $8$ is $1\cdot3\cdot(-1)^{997}\equiv-3\equiv5\pmod{8}$.\ldots Next, consider the product modulo $125$\ldots the product modulo $125$ is $9\cdot99\cdot(-1)^{997}\equiv-891\equiv16\cdot(-1)\equiv109\pmod{125}\ldots $Finally, by the Chinese Remainder Theorem, solving $x\equiv5\pmod{8}$ and $x\equiv109\pmod{125}$ yields $x=125m+109$ with $m\equiv0\pmod{8}$, hence $x\equiv109\pmod{1000}$. \textbf{\textcolor{OliveGreen}{Therefore, the remainder is boxed\{109\}}}.
\\
\textbf{Truncation Ratio:} 0.27

\\
\hline
\end{tabular}}

\caption{\textbf{Case Study.} Text in \textbf{\textcolor[rgb]{0.7,0.3,0.3}{red}} highlights erroneous parts in the reasoning and their underlying causes, text in \textbf{\textcolor{OliveGreen}{green}} denotes correct reasoning content, and text in \textbf{\textcolor{blue}{blue}} indicates the truncation point of the informative prefix.}
\label{case-study:method}
\end{table*}

\begin{table*}[t]
\centering
\small
\renewcommand\arraystretch{1.2}
\setlength{\fboxsep}{1.0pt}
\resizebox{\linewidth}{!}{ 
\begin{tabular}{p{0.88\textwidth}}
\hline
\rowcolor{gray!8}{\textbf{Case:} The Case of Outputs Under Different Distillation Methods.}

\\
\hline
\textbf{Question:} 
The roots of the polynomial $10x^3 - 39x^2 + 29x - 6$ are the height, length, and width of a rectangular box (right rectangular prism). A new rectangular box is formed by lengthening each edge of the original box by $2$. What is the volume of the new box?
\\
\textbf{Ground-Truth:} 30

\\
\textbf{Vanilla Response:} 
To solve the problem, we will follow these steps: first, find the roots of the polynomial $10x^3 - 39x^2 + 29x - 6$, then determine the dimensions of the original box, and finally compute the volume of the new box. 
Using the Rational Root Theorem, it is claimed that $x=2$ is a root, and the polynomial is factored as $(x-2)(10x^2-19x+3)$
\ldots
The resulting roots are 
$x=2$, $x=\frac{19+\sqrt{241}}{20}$, and $x=\frac{19-\sqrt{241}}{20}$. 
Each dimension is increased by $2$, leading to a new volume computed as 
$4 \cdot \frac{59+\sqrt{241}}{20} \cdot \frac{59-\sqrt{241}}{20} = 32.4$
\ldots
\textbf{\textcolor[rgb]{0.7,0.3,0.3}{Therefore, the final answer is boxed\{32.4\}}}.
\\

\textbf{SFT (Label):} 
 \textbf{\textcolor[rgb]{0.7,0.3,0.3}{Thus, the volume of the new box is \{37.5\}}}.

\\
\textbf{SFT (Long-CoT):} 
Alright, so I have this problem here where I need to find the volume of a new rectangular box formed by increasing each edge of an original box by $2$ units.
The original box's dimensions are the roots of the polynomial $10x^3 - 39x^2 + 29x - 6$.
Hmm, okay. Let me break this down step by step \ldots \textbf{\textcolor{OliveGreen}{That reminds me of Vieta’s formulas.
For $10x^3 - 39x^2 + 29x - 6$, we have
$h+l+w=\tfrac{39}{10}$,
$hl+hw+lw=\tfrac{29}{10}$,
and $hlw=\tfrac{3}{5}$}}\ldots Now expand the volume:
$(h+2)(l+2)(w+2)=hlw+2(hl+hw+lw)+4(h+l+w)+8$

\ldots

At this point, I want to double-check the arithmetic to avoid mistakes. Converting everything to a common denominator,
I rewrite $8$ as $\tfrac{80}{10}$ and sum the terms,
\textbf{\textcolor[rgb]{0.7,0.3,0.3}{obtaining $\tfrac{58+156+80}{10}$}}\ldots
Wait, let me verify this again to be safe.\ldots After rechecking, the total is interpreted as $\tfrac{294}{10}$,
which simplifies to $29.4$.
The steps seem consistent after revision\ldots
\textbf{\textcolor[rgb]{0.7,0.3,0.3}{So the final answer is boxed\{29.4\}}}.

\\
\textbf{\method{}:} 
\texttt{<Begin\_of\_prefix>}
Okay, so I need to find the volume of a new rectangular box formed by increasing each edge of an original box by $2$ units. 
The original box's dimensions (length, width, height) are the roots of the polynomial $10x^3 - 39x^2 + 29x - 6$. 
First, I need to find those roots, or at least use their relationships. 
\textbf{\textcolor{OliveGreen}{I remember Vieta's formulas: for a cubic equation $ax^3+bx^2+cx+d=0$, the sum of the roots is $-b/a$, the sum of pairwise products is $c/a$, and the product of the roots is $-d/a$}}\ldots \texttt{<End\_of\_prefix>}\\

To continue solving the question based on the  existing prior knowledge \ldots 
Therefore, the roots $(L,W,H)$ satisfy 
$L+W+H=\frac{39}{10}$, $LW+WH+HL=\frac{29}{10}$, and $LWH=\frac{3}{5}$. 
The new volume is $(L+2)(W+2)(H+2)$, which expands to 
$LWH + 2(LW+WH+HL) + 4(L+W+H) + 8$. 
Substituting the values gives 
$\frac{3}{5} + 2\cdot\frac{29}{10} + 4\cdot\frac{39}{10} + 8 = 30$. 
\ldots
\textbf{\textcolor{OliveGreen}{Therefore, the new volume is boxed\{30\}}}.
\\
\hline
\end{tabular}}

\caption{Case Study. Text in \textbf{\textcolor[rgb]{0.7,0.3,0.3}{red}} highlights erroneous parts in the reasoning and their underlying causes, text in \textbf{\textcolor{OliveGreen}{green}} denotes correct reasoning content.}
\label{case-study:different_strategy}
\end{table*}

\end{document}